\pdfoutput=1

\documentclass[11pt]{article}

\usepackage[]{ACL2023}

\usepackage{times}
\usepackage{latexsym}

\usepackage[T1]{fontenc}

\usepackage[utf8]{inputenc}

\usepackage{microtype}

\usepackage{inconsolata}
\usepackage[color=red]{todonotes}
\usepackage{arydshln}
\usepackage{subcaption}
\usepackage{multirow}
\usepackage{comment}
\usepackage{booktabs}
\usepackage{CJKutf8}

\renewcommand{\sectionautorefname}{\S\kern-0.2em}
\renewcommand{\subsectionautorefname}{\S\kern-0.2em}
\renewcommand{\subsubsectionautorefname}{\S\kern-0.2em}

\title{How Good Are GPT Models at Machine Translation?\\ A Comprehensive  Evaluation}

\author{Amr Hendy,  Mohamed Abdelrehim,  Amr Sharaf,  Vikas Raunak, \\{\bf Mohamed Gabr, Hitokazu Matsushita,  Young Jin Kim,}\\ {\bf Mohamed Afify, Hany Hassan Awadalla\thanks{Corresponding author: hanyh@microsoft.com}}
  \\
  {\bf Microsoft} \\}

\begin{document}
\maketitle
\begin{abstract}

Generative Pre-trained Transformer (GPT) models have shown remarkable capabilities for natural language generation, but their performance for machine translation has not been thoroughly investigated. In this paper, we present a comprehensive evaluation of GPT models for machine translation, covering various aspects such as quality of different GPT models  in comparison with state-of-the-art research and commercial systems, effect of prompting strategies, robustness towards domain shifts  and document-level translation. We experiment with eighteen different translation directions involving high and low resource languages, as well as non English-centric translations, and evaluate the performance of three GPT models: ChatGPT, GPT3.5 (text-davinci-003), and text-davinci-002. Our results show that GPT models achieve very competitive translation quality for high resource languages, while having limited capabilities for low resource languages. We also show that hybrid approaches, which combine GPT models with other translation systems, can further enhance the translation quality. We perform comprehensive analysis and human evaluation to further understand the characteristics of GPT translations. We hope that our paper provides valuable insights for researchers and practitioners in the field and helps to better understand the potential and limitations of GPT models for translation.

\end{abstract}

\section{Introduction}

 Recent advancements in natural language processing (NLP), particularly the development of large-scale language modeling techniques, have brought remarkable improvements in machine translation as well as in other NLP tasks~\citep{fan2021beyond, kim2021scalable, costa2022no}. The emergence of large language models with diverse capabilities, including machine translation, has opened up new possibilities for building more effective translation systems~\citep{gpt-3, chowdhery2022palm}. Among these models, the latest Generative Pre-trained Transformer (GPT) models~\citep{gpt-3} have gained significant attention for their ability to generate coherent and context-aware text. We present a comprehensive evaluation of GPT models for machine translation, exploring their strengths and limitations, and providing insights for researchers and practitioners working in the area of machine translation.


GPT models and the conventional Neural Machine Translation (NMT) systems are both based on the transformer architecture~\cite{vaswani2017}, but they differ in several aspects. First, GPT models are decoder-only models that use the same parameters to process the context and the source as a single input for generating the next output. On the other hand, NMT models usually have an encoder-decoder architecture that encodes the source sentence in the encoder network and decodes the target sentence conditioned on the previous outputs in the decoder network. Second, GPT models are mainly trained on monolingual data, with a strong bias towards English\footnote{\url{https://github.com/openai/gpt-3/blob/master/dataset_statistics/languages_by_character_count.csv}}, whereas NMT models rely on large amounts of  highly curated parallel data. Third, GPT models need a much larger number of parameters to achieve multilingual in-context capabilities. We have observed that GPT models exhibit promising translation capabilities, even with these differences in architecture and training data.


The performance of GPT models in machine translation, despite their promising potential, remains under-investigated relative to commercial and state-of-the-art research systems. This study aims to address this research gap by systematically assessing the efficacy of GPT models for machine translation, with a focus on their performance, prompts, document-level translation, domain robustness and possible advantages of integrating them with conventional NMT systems.

To explore the potential of GPT models for translation, we perform comprehensive experiments to examine their translation abilities. Specifically, we investigate the performance of GPT models for machine translation across 18 language pairs, covering both high and low-resource languages, as well as English-centric and non-English-centric directions. We compare the quality of three GPT models: \emph{text-davinci-002}, \emph{text-davinci-003} (GPT3.5), and \emph{ChatGPT}, and show that they differ significantly in their translation capabilities.

We also explore the impact of prompting strategies on the performance of GPT models for machine translation. We examine both the content and the form of the prompts, and identify the best practices for obtaining optimal results. Furthermore, we test the hypothesis that GPT models would enhance document-level translation, as they could exploit the context and coherence of the entire document to generate more accurate and fluent translations. We evaluate this hypothesis on various language pairs using several metrics. Additionally, we evaluate the cross-domain generalization ability of GPT models for translation tasks and examine their robustness under domain shift.

Moreover, we conduct extensive human evaluations and analyses to provide valuable insights into the strengths and weaknesses of GPT models for machine translation, and to suggest directions for future work. We also perform comprehensive analyses to understand whether GPT and NMT models have complementary characteristics, and we propose several ideas to combine the advantages of the two paradigms. Finally, we touch on the effectiveness of GPT models on cross-lingual natural language tasks beyond translation, and explore their multilingual capabilities and limitations.

To explore the research questions above, we organize the paper as follows:

\begin{itemize}
\item We provide a detailed experimental setup (\autoref{sec:exp-setup}), which includes the datasets used (\autoref{sec:datasets}), the machine translation systems used in the comparison(\autoref{sec:systems}), the GPT systems (\autoref{sec:gpt}), and the evaluation methods (\autoref{eval}).
\item We present a series of experiments (\autoref{sec:results}) that investigate different aspects of GPT models for machine translation. These experiments cover prompt selection strategies (\autoref{sec:prompt}), zero-shot translation capabilities of GPT models~(\autoref{sec:zero-shot}), GPT performance on high-resource languages (\autoref{sec:high-resource}), GPT performance on low-resource and non-English-centric languages (\autoref{sec:low-resource}), document-level MT with GPT (\autoref{sec:document}), translation robustness toward domain shift (\autoref{sec:domains}), and hybrid GPT and NMT translation (\autoref{sec:hybrid}).
\item We then present the human evaluation and analysis that provides insights into the quality of GPT translations (\autoref{sec:humaneval}).
\item We discuss the characteristics of GPT translations (\autoref{sec:analysis}) vis-à-vis NMT and analyze the differentiating aspects of GPT translations (\autoref{sec:analysis1}) by quantitatively enumerating language modeling bias artifacts (\autoref{sec:bias}), the characteristics of translation across various language directions (\autoref{sec:xe}-\autoref{sec:xy}), as well as parallel data bias artifacts (\autoref{sec:parallel}).
\item We explore the multilingual capabilities of GPT models beyond translation (\autoref{sec:beyond}).
\item We conclude by summarizing our findings and suggesting future directions for research (\autoref{sec:conclusion}).
\end{itemize}

\section{Experimental Setup}
\label{sec:exp-setup}


\subsection{Datasets}
\label{sec:datasets}

We considered 18 different translation directions across a diverse set of languages for our comprehensive evaluation. The evaluation covers both high- and low-resource languages, as well as English-centric and non-English-centric direct translations. The languages considered in this study include European (English-\textit{EN}, French-\textit{FR}, German-\textit{DE}, Czech-\textit{CS} and  Icelandic-\textit{IS}), Asian (Chinese-\textit{ZH} and Japanese-\textit{JA}), Cyrillic (Russian-\textit{RU} and Ukrainian-\textit{UK})  and African (Hausa-\textit{HA}). 

We use publicly available datasets to facilitate reproducibility and data sharing. We use the WMT22 testsets~\footnote{\url{https://www.statmt.org/wmt22/translation-task.html}} for all languages except for Icelandic and Hausa, for which we use the WMT21 testsets~\footnote{\url{https://www.statmt.org/wmt21/translation-task.html}}. We use WMT22 datasets for two reasons: first, they are recent and less likely to overlap with the GPT models training data that was collected until June 2021 \footnote{\url{https://help.openai.com/en/articles/6639781-do-the-openai-api-models-have\\-knowledge-of-current-events}}. Second, they have natural source texts and translated target texts, which avoid the problems of using “translationese” testsets with unnatural source texts in the original language that may  result in inaccurate evaluation~\cite{zhang-toral-2019-effect}.

\autoref{tab:datasets} summarizes the datasets and sizes used in this paper.  We focus on the most recent datasets with directional translation, as older datasets may have influenced the training data of the GPT models. We make all data and analysis in this paper publicly available to promote further research.~\footnote{\url{https://github.com/microsoft/gpt-MT}}.


\begin{table}[!htbp]
\centering
\scalebox{0.80}{
\begin{tabular}{c|c|c|c}
\hline
\multirow{2}{*}{\textbf{Lang-Pair}} & \multirow{2}{*}{\textbf{Dataset}} & \textbf{Number of}  & \multirow{2}{*}{\textbf{WMT-Best System}} \\
 &  & \textbf{sentences}  &  \\ \hline
CS-EN & \multirow{2}{*}{WMT22}  & 1448  & Online-W \\ 
EN-CS &   & 2037  & Online-W \\ \hline
DE-EN & \multirow{2}{*}{WMT22}  & 1984  & Lan-Bridge \\ 
EN-DE &   & 2037  & Online-B \\  \hline
IS-EN & \multirow{2}{*}{WMT21} & 1000 & Facebook-AI \\
EN-IS &  & 1000 & Facebook-AI\\  \hline
JA-EN & \multirow{2}{*}{WMT22}  & 2008 & DLUT \\ 
EN-JA &   & 2037 & NT5 \\   \hline
ZH-EN & \multirow{2}{*}{WMT22} & 1875 & JDExploreAcademy \\
EN-ZH &  & 2037 & Online-W \\  \hline
UK-EN & \multirow{2}{*}{WMT22}  & 2018  & Lan-Bridge \\ 
EN-UK &   & 2037  & Online-B \\   \hline
RU-EN & \multirow{2}{*}{WMT22} & 2016 & JDExploreAcademy \\
EN-RU &  & 2037 & Online-W \\  \hline
HA-EN & \multirow{2}{*}{WMT21} & 997 & Facebook-AI \\
EN-HA &  & 1000 & Facebook-AI \\  \hline
FR-DE & \multirow{2}{*}{WMT22} & 2006 &  Online-W \\
DE-FR &  & 1984 &  Online-B \\ 
\hline
\end{tabular}}
\caption{Test datasets used in the evaluation and best systems used for comparison as reported in WMT by~\citet{kocmi-EtAl:2022:WMT} and \citet{wmt-2021-machine}.}
\label{tab:datasets}
\vspace{-1.5em}
\end{table}


\subsection{Neural Machine Translation Systems}
\label{sec:systems}

In this study, we compare the performance of GPT systems against both state-of-the-art (SoTA) research  and commercial systems. We use the  top ranked systems (WMT-Best) in the WMT evaluation campaigns for each language pair, which we use as a baseline for comparison. WMT-Best systems are a mix of top ranked commercial and research systems. We use the system outputs as provided by the evaluation campaigns \cite{kocmi-EtAl:2022:WMT, wmt-2021-machine}. \autoref{tab:datasets}  shows the list of top ranked system for each language pair. We also utilize \texttt{Microsoft Translator} which we access through the public API available on \texttt{Azure Cognitive Services}~\footnote{\url{https://azure.microsoft.com/en-us/products/cognitive-services/translator}}.

\subsection{GPT Systems}
\label{sec:gpt}

We assess the latest three variants of the largest GPT models available, as listed at OpenAI's documentation\footnote{\url{https://beta.openai.com/docs/model-index-for-researchers}}. These models are:
\begin{itemize}
\item \emph{text-davinci-002} - an InstructGPT model ~\cite{chung-etal-2022-instructgpt} which utilizes Reinforcement Learning with reward models trained based on human comparisons.
\item \emph{text-davinci-003} - an improved version of \emph{text-davinci-002}.
\item \emph{ChatGPT} - a model that is similar to the previous two and optimized specifically for conversational purposes\footnote{\url{https://openai.com/blog/chatgpt}}.
\end{itemize}

All GPT models have been accessed through APIs on \texttt{Microsoft Azure OpenAI Service}~\footnote{\url{https://azure.microsoft.com/en-us/products/cognitive-services/openai-service}}.

\subsection{Evaluation Methods}
\label{eval}
 \paragraph{Sentence-Level Evaluation} 

The MT Metrics shared task \cite{metrics22} recommends the use of neural network-based metrics in machine translation evaluation, as they have demonstrated a high correlation with human evaluation and are resilient to domain shift. Following these recommendations, we employ the top-ranked metrics from \textit{Unbabel} in the shared task~\footnote{\url{https://github.com/Unbabel/COMET}}. Specifically, we use COMET-22 (\textit{wmt22-COMET-da}) \cite{COMET-22}, a reference-based metric that combines direct assessments (DA), sentence-level scores, and word-level tags from Multidimensional Quality Metrics (MQM) error annotations. For reference-less quality estimation, we adopt COMETkiwi (\textit{wmt22-COMETkiwi-da}) \cite{comet-kiwi}. Additionally, we report results using \textit{SacreBLEU}\footnote{\url{https://github.com/mjpost/sacrebleu}} and \textit{Chrf}\cite{popovic-2015-chrf} for completeness, although we note that these metrics are not extensively used in our analysis.


 \paragraph{Document-Level Evaluation}
 For the experiments on document-level translation using GPT, we face a challenge in evaluating performance due to the lack of metrics that can handle the non one-to-one sentence mapping that the systems may produce. To address this challenge, we have adapted the COMET metrics to better suit document-level evaluation. The adaptation involves splitting the document into multiple segments with an overlapping sliding window and computing the average score across these segments to compare two documents. We use Doc-COMETkiwi to refer to the modified metric throughout the evaluation.
 
This simple modification has three clear design advantages over a pure sentence-level evaluation. First, it allows each sentence to be evaluated within its context. Second, it enables each sentence to be evaluated across multiple contexts due to the overlapping nature of the sliding window. Lastly, it avoids the limited context window of the evaluation models that could hinder quality assessment over longer static windows on documents. We do not claim that this is an optimal metric for evaluating document-level translation, but it overcomes the limitation of the one-to-one sentence mapping and may capture the quality of translation in ambiguous contexts better than sentence-level representation. We argue that developing more robust document-level metrics is still essential.

The current metrics for evaluating machine translation performance may not be adequate for measuring the performance of GPT models, and it may be necessary to develop new metrics that take into account the unique characteristics of these models.

\paragraph{Human Evaluation and Analysis}
We perform human evaluation (\autoref{sec:humaneval}) using source-based sentence-level contrastive Direct Assessment + Scalar Quality Metric (contrastive DA+SQM; \citealt{akhbardeh2021findings}, \citealt{kocmi2022findings}), with annotations provided by professional annotators. We also conduct thorough analysis on various characteristics of the translation (\autoref{sec:analysis}).

\section{Experiments}
\label{sec:results}

In this section, we present various experiments. In \autoref{sec:prompt}, we describe several prompt selection strategies. In \autoref{sec:zero-shot}, we evaluate various GPT models in a zero-shot setup. In \autoref{sec:high-resource}, we show results for high-resource language pairs, followed by results for low-resource and non-English pairs in \autoref{sec:low-resource}. \autoref{sec:document} provides document translation results. \autoref{sec:domains} examines the robustness of GPT models under domain shift. Finally, in \autoref{sec:hybrid}, we discuss the potential of combining the benefits of GPT and NMT models.


\subsection{Prompt Selection Strategies}
\label{sec:prompt}
It has been shown that the performance of LLMs can be enhanced through in-context learning by providing few labelled examples (prompts) in addition to the test input \cite{gpt-3}. This few-shot paradigm has demonstrated strong performance across multiple natural language processing (NLP) tasks \cite{chung-etal-2022-instructgpt, goyal2022news, wei2022chain, chowdhery2022palm}. There has also been a series of recent works on in-context learning for machine translation (MT) with rather mixed results and various ways of shot selection. Recently, \citet{Zhang-etal-2023-promptingLMs} use GLM-130B and show consistent but rather low correlation between the performance of MT and shot selection as compared to random. They use different features that show varying level of correlation to the performance. In the same spirit, \citet{vilar-etal-2022-googleincontext} use different prompt selection schemes with PaLM-540B model with the main conclusion that shot selection is not necessarily better than random but they point out the importance of using high quality shots. In the same vein, \citet{Agrawal-etal-2022-fbincontext} use a much smaller model XGLM-7.5B and multiple selection criteria for the shots. They show that a combination of a retrieval and task metrics are consistently better than the random baseline across different translation directions.   



In this paper, we explore prompt selection strategies along two dimensions: quality and relevance. Our pool to select few-shot examples is the cleaned WMT training data for each direction. This is obtained by filtering the full training data using language identification  and length ratio. The size of the full and cleaned training data is shown in \autoref{tab:datasets-train} for each direction. We do not use the development data from WMT shared tasks, despite its high quality, for shot selection to avoid any small chance of leaking information about the test set. In all cases, we test the performance with 0, 1 and 5 shots. In our preliminary experiments, we found that increasing beyond 5 shots did not result in any meaningful improvement. We show below how we select shots based on quality and relevance.

\begin{itemize}
\item \emph{Quality}: To ensure high quality shots we sort our training data using LaBSE \cite{feng-etal-2020-labse}. We consider high quality shots that are randomly chosen from the top 1 million pairs as opposed to the full data. We also found it useful to select sentences that are longer than 50 tokens. 
\item  \emph{Relevance}: We consider relevant shots that are close to the input sentence. Based on preliminary experiments\footnote{We also tried using COMET and embeddings based on a trained NMT model.} we use the cosine distance between LaBSE embeddings as a measure of closeness. We always select relevant pairs from high quality ones (the top 1M pairs from LaBSE-scored training data). For computational efficiency, we adopt a two stage approach. We first use the input text and apply elastic search\footnote{Based on BM25 text retrieval.} to retrieve the top 64 pairs then return the top 1 or 5 shots based on the LaBSE distance.  
\end{itemize}
In the results, we refer to full random as \emph{RR} (Random)  while high quality are referred to as \emph{QR} (Quality Random). The high quality shots selected  through relevance are referred to as \emph{QS} (Quality Selected).

\subsection{Zero-Shot Translation Capabilities of GPT Models}
\label{sec:zero-shot}
We compare the general zero-shot translation capabilities of the three GPT models on four language pairs, in eight distinct translation directions. The selected languages were chosen with a focus on balancing representation. The languages include 1) German, which is one of the most  represented non-English languages in GPT training data, 2) Russian, a large-scale non-Latin language, 3) Chinese, which represents a large-scale language with a script distinct from that of the majority of training data languages, and 4) French-German pair  as non English-centric use case.

\begin{table*}[!htbp]
\centering
\small
\vspace{-0.8em}
\scalebox{1.0}{
\begin{tabular}{c|c c c c | c c c c}
\hline
\textbf{System} &  \textbf{COMET-22}  & \textbf{COMETkiwi} & \textbf{ChrF}  & \textbf{BLEU} &
\textbf{COMET-22}  & \textbf{COMETkiwi} & \textbf{ChrF}  & \textbf{BLEU} \\ \hline
& \multicolumn{4}{c}{DE-EN} &   \multicolumn{4}{c}{EN-DE} \\ 
WMT-Best & \textbf{85.0} &	\textbf{81.4} &	\textbf{58.5} &	\textbf{33.4} & \textbf{87.2} & \textbf{83.6} &	\textbf{64.6} &	\textbf{38.4}\\\hdashline
text-davinci-002 & 73.2	 & 73.1	& 46.1 & 	23.3 & 82.0	& 79.0 & 56.0 &28.6\\
text-davinci-003 & 84.8$^*$	&81.2$^*$	&56.8 &	30.9 &85.6$^*$	&82.8$^*$	&60.2$^*$	&31.8$^*$\\
ChatGPT & {84.8}$^*$&	{81.1} &	{58.3}$^*$&	{33.4}$^*$ & 84.2	&81.0&	59.6&	30.9 \\\hline

& \multicolumn{4}{c}{ZH-EN} &   \multicolumn{4}{c}{EN-ZH} \\
WMT-Best & 81.0	& 77.7 & 	\textbf{61.1} &	\textbf{33.5} & \textbf{86.7}	& \textbf{82.0}	 & \textbf{41.1} &	\textbf{44.8}\\\hdashline
text-davinci-002 & 74.1 & 	73.1 & 	49.6 & 	20.6 & 84.0 & 	79.0 & 	32.1 & 36.4 \\
text-davinci-003 &\textbf{81.6}$^*$	&\textbf{78.9}$^*$	&56.0$^*$ &	25.0 & 85.8$^*$	& 81.3$^*$	&34.6&	38.3\\
ChatGPT & 81.2	&78.3&	56.0	&25.9$^*$& 84.4&	78.7&	36.0$^*$&	40.3$^*$\\\hline

& \multicolumn{4}{c}{RU-EN} &   \multicolumn{4}{c}{EN-RU} \\
WMT-Best &\textbf{86.0} & 	\textbf{81.7} &	\textbf{68.9} & 	\textbf{45.1} &  \textbf{89.5} &	\textbf{84.4} &	\textbf{58.3}	& \textbf{32.4}\\\hdashline
text-davinci-002 & 77.5 & 76 & 58.7 & 34.9 & 85.4 & 80.9 & 	51.6 & 	25.1\\
text-davinci-003 &84.8$^*$ &81.1$^*$&64.6	&38.5 &86.7$^*$&	82.2$^*$&	54.0$^*$&	27.5$^*$\\
ChatGPT & 84.8$^*$	&81.0&	66.5$^*$&	41.0$^*$ &77.6	&70.4&	41.1&	19.0\\\hline

& \multicolumn{4}{c}{FR-DE} &   \multicolumn{4}{c}{DE-FR} \\
WMT-Best & \textbf{89.5}	& \textbf{80.7} &	\textbf{81.2}&	\textbf{64.8} & \textbf{85.7}	& 79.5 &	\textbf{74.6}&	\textbf{58.4}\\\hdashline
text-davinci-002 & 66.6 & 67.9 & 	45.8 & 	25.9 & 64.2 & 	67.6 & 	44.6 & 	24.5\\
text-davinci-003 & 84.6&	77.9&	65.7$^*$&	42.5$^*$& 78.5	&76.1&	58.9&	35.6\\
ChatGPT &84.7$^*$	&78.5$^*$&	65.2&	42.0 &81.6$^*$	&\textbf{79.8}$^*$&	60.7$^*$	&37.3$^*$ \\\hline

\end{tabular}}
\caption{Zero-Shot evaluation results with three GPT models on 8 language pairs from WMT22 Testset. The best scores across different systems are marked bold. * denotes the best results among GPT systems.}
\label{tab:result.zs-wmt}
\vspace{-1.5em}
\end{table*}

In this experiment, we compare the performance of three GPT models \emph{text-davinci-002}, \emph{text-davinci-003}, and \emph{ChatGPT} with the top ranked systems in WMT22, as shown in \autoref{tab:result.zs-wmt}. Remarkably, \emph{text-davinci-002} shows lower performance, underperforming across all language pairs compared to the other two GPT models. On the other hand, \emph{text-davinci-003} clearly shows better translation performance across all languages in this evaluation. Its zero-shot performance is comparable to the best performing WMT DE-EN system and outperforms the best ZH-EN WMT system.

ChatGPT shows a storng performance in the DE-EN language pair, while performing similarly to \emph{text-davinci-003} when translating to English as well as French-German pairs. In terms of translating from English to other languages, \emph{text-davinci-003} shows better performance than the other two GPT models. It is noteworthy that the translation between French and German, which is not English-centric, exhibits surprising competitiveness with state-of-the-art systems, despite the fact that the majority of the training data used for the GPT model is English-centric.

While both COMETkiwi and COMET-22 show relevant results, both lexical metrics (BLEU and ChrF) show consistent degradation with GPT models. This is consistent with similar findings from \cite{vilar-etal-2022-googleincontext} on PALM-540B model. We conducted human evaluation and more thorough analysis to further understand such results.

From these results, we can see that the three variants of GPT models exhibit different characteristics. However, the nature and extent of these differences remain unclear and require further investigation, depending on the availability of more information regarding the models, their training data, and their training methods. This superior performance of \emph{text-davinci-003}, which is achieved in a zero-shot setting, motivates further investigation into the effect of few-shot in-context learning and shot selection strategies. We investigate these questions further in the sections below.

\subsection{GPT Performance on High-resource Languages}
\label{sec:high-resource}

Given the results in the previous section, we focus on evaluating  \emph{text-davinci-003} model, expanding the scope of the study to 18 language pairs and comparing its performance with that of a commercial system (Microsoft Translator) in addition to WMT SoTA systems. For consistency, in all the subsequent results, we use the term “GPT” to denote the \emph{text-davinci-003} model, unless explicitly stated otherwise. 

We experiment  with  various shot selection strategies:  zero-shot, random (RR), quality (QR) and relevance selected (QS) prompts as described in~\autoref{sec:prompt}. We report results for 1 and 5 shots along with the best WMT systems and MS-Translator.

\begin{table*}[!htbp]
\centering
\vspace{-0.8em}
\scalebox{0.9}{
\begin{tabular}{c|c c c c | c c c c}
\hline
\textbf{System} &  \textbf{COMET-22}  & \textbf{COMETkiwi} & \textbf{ChrF}  & \textbf{BLEU} &
\textbf{COMET-22}  & \textbf{COMETkiwi} & \textbf{ChrF}  & \textbf{BLEU} \\ \hline
& \multicolumn{4}{c}{DE-EN} &   \multicolumn{4}{c}{EN-DE} \\ 

WMT-Best & 85.0 &	81.4 &	58.5 &	33.4 & \textbf{87.2} & \textbf{83.6} &	\textbf{64.6} &	\textbf{38.4} \\
MS-Translator &84.7&	81.0&	 \textbf{58.5}&	\textbf{33.5} & 86.8&	83.4&	64.2&	37.3 \\\hdashline
GPT  Zeroshot & 84.8	&81.2	&56.8 &	30.9 &85.6	&82.8	&60.2	&31.8\\
GPT 1-Shot RR & 84.9&	81.3&	56.1&	30.4& 86.1&	83.0&	60.7&	31.9\\
GPT 1-Shot QR &  84.9&	81.3&	56.7&	31.1 &85.8	& 82.8&	60.7&	32.4\\
GPT 5-Shot RR &  85.2&	81.5&	56.5&	31.2& 86.5$^*$ &	83.2 $^*$&	61.0&	32.4 \\
GPT 5-Shot QR & \textbf{85.4}$^*$	&  \textbf{81.5}$^*$ &	57.7&	32.4 &86.4&	83.1&	61.3$^*$&	33.2$^*$ \\ \hdashline[0.5pt/5pt]
GPT 5-Shot QS 	& 85.0	& 81.3	& 57.8$^*$	& 32.5$^*$ & 85.9 &	82.9&	60.8 &	32.7 \\ \hline

& \multicolumn{4}{c}{CS-EN} &   \multicolumn{4}{c}{EN-CS} \\ 
WMT-Best & \textbf{89.0} &	\textbf{82.5} &	\textbf{79.3} &	\textbf{64.2} & \textbf{91.9} &	\textbf{85.3} &	\textbf{68.2} &	\textbf{45.8}\\
MS-Translator & 87.4	& 82.2&	74.0	& 54.9 & 90.6	& 84.2	& 65.6	& 42.1  \\ \hdashline
GPT  Zeroshot & 86.2	&82.0	&67.5&	44.5& 88.6&	82.9	&57.9	&31.3\\
GPT 1-Shot RR &  86.6	&82.3	&67.9	&45.4& 89.7$^*$	&84.0$^*$ &	58.3&	31.6\\
GPT 1-Shot QR & 86.4	&82.3	&67.8&	45.0& 89.2&	83.6&	58.6&	32.5 \\
GPT 5-Shot RR & 86.6&	82.3&	66.4&	44.2& 89.4&	83.8	&58.6&	32.0 \\\
GPT 5-Shot QR & 86.9$^*$ &	\textbf{82.5}$^*$&	69.2$^*$&	47.5$^*$ &89.0&	83.3&	59.0$^*$& 32.9$^*$ \\\hline

& \multicolumn{4}{c}{JA-EN} &   \multicolumn{4}{c}{EN-JA} \\ 
WMT-Best & 81.6 &	80.3 &	\textbf{49.8} &	\textbf{24.8} & 	\textbf{89.3}	 & 	\textbf{85.8}	 & 	\textbf{36.8} &		\textbf{27.6}\\
MS-Translator &  81.5	& 80.1 & 49.6 &	24.5 & 88.0	&85.3&	34.9&	25.1\\ \hdashline
GPT  Zeroshot & 81.5 &	80.7 &	47.7&	21.1 & 87.8 &	84.8 &	31.2 &	21.2\\
GPT 1-Shot RR & 81.7&	80.7&	46.8&	20.2& 88.3&	85.1&	31.8&	22.0 \\
GPT 1-Shot QR & 81.6 &	80.8 &	48.3$^*$ &	22.1 & 88.4$^*$ &	85.3 &	32.2$^*$ &	22.5$^*$ \\
GPT 5-Shot RR & \textbf{82.0}$^*$ &	\textbf{80.9}$^*$ &	48.2 &	22.4$^*$ &88.2 &	85.4$^*$	&31.7	&21.4 \\
GPT 5-Shot QR & 81.8	& 80.8	&47.2 &	21.0 &  88.2 &	85.3 &	31.1 &	21.6 \\\hline

& \multicolumn{4}{c}{ZH-EN} &   \multicolumn{4}{c}{EN-ZH} \\
WMT-Best & 81.0	& 77.7 &  \textbf{61.1} & \textbf{33.5} & 86.7	& 82.0	 & 41.1 &	44.8\\
MS-Translator &80.4& 77.6	& 57.7 & 	27.9 & 86.1 & 	81.4	& \textbf{43.1} & 	\textbf{48.1}\\ \hdashline
GPT  Zeroshot & 	\textbf{81.6}$^*$	& \textbf{78.9}$^*$	&56.0$^*$ &	25.0$^*$ & 85.8	&81.3	&34.6&	38.3\\
GPT 1-Shot RR & 80.9&	78.2&	55.2&	24.2& 86.7&	81.8&	38.7&	42.8 \\
GPT 1-Shot QR & 81.2 &	78.8 &	55.3	& 24.2 & 86.1 &	81.5 &	35.5 &	38.8 \\
GPT 5-Shot RR & 81.1	& 78.8 &	55.0 &	24.4 & 87.0 &	82.0&	37.1&	41.3 \\\
GPT 5-Shot QR &81.1	& 78.7 &	54.7 &	23.8 & \textbf{87.0}$^*$ & 	\textbf{82.2}$^*$ &	39.8$^*$ &	43.7$^*$ \\ \hdashline[0.5pt/5pt]
GPT 5-Shot QS & 81.0 &	78.5 &	55.5 &	24.6 & 86.2	& 81.5	& 38.3	 & 41.8 \\\hline

& \multicolumn{4}{c}{RU-EN} &   \multicolumn{4}{c}{EN-RU} \\
WMT-Best &\textbf{86.0} & 	\textbf{81.7} &	\textbf{68.9} & 	\textbf{45.1} &  \textbf{89.5} &	\textbf{84.4}&	\textbf{58.3}	& 32.4 \\
MS-Translator &85.2&	80.7&	68.3&	43.9& 87.4&	82.9&	58.1&	\textbf{33.1}\\ \hdashline
GPT  Zeroshot &84.8&	81.1&	64.6	&38.5 &86.7&	82.2&	54.0&	27.5\\
GPT 1-Shot RR &84.1&	80.6&	63.3&	37.9 &86.4&	81.9	&54.3	&28.1 \\
GPT 1-Shot QR & 84.9 &	81.2$^*$ &	65.4$^*$	& 40.1 & 86.9 &	82.4	& 53.8	& 27.5 \\
GPT 5-Shot RR &84.9 &	81.2$^*$ &	63.9 &	39.0 &  86.8 &	82.3& 54.3 &	27.9 \\\
GPT 5-Shot QR & 84.9 &	81.0 &	65.4$^*$ &	40.0 &  87.0$^*$ &	82.4$^*$ &	54.4$^*$ &	28.2$^*$ \\  \hdashline[0.5pt/5pt]
GPT 5-Shot QS & 85.0$^*$ &	81.2$^*$	& 65.3 &	40.2$^*$ & 86.4	& 82.2	& 54.4$^*$	& 28.0 \\\hline

\hline

& \multicolumn{4}{c}{UK-EN} &   \multicolumn{4}{c}{EN-UK} \\
WMT-Best & \textbf{86.0}	& \textbf{81.5} &	\textbf{67.3} &	 \textbf{44.6} & \textbf{88.8}	& \textbf{83.4} &	\textbf{59.3}	& \textbf{32.5}\\
MS-Translator & 83.5&	79.7	&65.3&	42.4& 86.1	&81.9	&56.1	&28.2\\\hdashline
GPT  Zeroshot &83.5	&80.1&	59.8&	34.8 &83.7 &	79.5 &	49.6	&21.1\\
GPT 1-Shot RR & 83.5&	80.3	&60.3&	35.6& 84.7&	80.2&	50.1&	21.2 \\
GPT 1-Shot QR &  83.8 &	80.3 &	61.4 &	37.5 & 85.1	 & 80.5 & 50.5	& 21.9 \\
GPT 5-Shot RR &83.6&	80.3&	58.8&	34.4& 85.4$^*$ &	80.8$^*$	&50.9$^*$&	22.6$^*$ \\\
GPT 5-Shot QR & 83.9$^*$ &	80.3$^*$ &	62.1$^*$ &	38.4$^*$  & 85.4 &	80.6 &	50.6 &	22.1 \\\hline

\end{tabular}}
\caption{Zero-Shot  and Few-Shots evaluteion results with GPT(text-davinci-003) on high resource languages from WMT Testsets. The best scores across different systems are marked bold. * denotes the best results among GPT systems.}
\label{tab:result.few-full-wmt}
\vspace{-1.5em}
\end{table*}

\autoref{tab:result.few-full-wmt} shows the performance of GPT \emph{text-davinci-003} with few-shot setups on high-resource languages from WMT Testsets.  With both reference and reference-less COMET scores, the model achieved impressive zero-shot results for all languages when translating into English. However, the few-shot configurations did not yield significant improvements over the zero-shot setup. GPT surpassed both the WMT-Best  and MS-Translator systems for DE-EN, JA-EN and ZH-EN language pairs, and almost matched the best systems for the other three language pairs. On the other hand, when translating from English to other languages, the few-shot setup consistently improved over the zero-shot setup, with most gains obtained from a single high-quality shot.  GPT outperformed both the WMT-Best and MS-Translator systems for EN-JA and EN-ZH language pairs. We experimented with high quality shots with relevance scores (QS) for three languages (German, Russian and Chinese), but we observed no improvements over quality shots alone. This result emphasises the importance of fewer quality shots especially when translating from English. This difference in behavior is consistent with the observations that the critical role of demonstrations within in-context learning is to provide specifications of the output space \cite{min2022rethinking, anonymous2023dissecting}, with a denser in-context learning signal being preferred when translating from English to other languages.

Similar to the  Zero-Shot results form \autoref{sec:zero-shot}, we observe that the lexical metrics are showing consistent degradation with all GPT models and configurations.


\subsection{GPT Performance on Low-resource  and non English-centric Languages}
\label{sec:low-resource}
\begin{table*}[!htbp]
\small
\centering
\vspace{-0.8em}
\scalebox{1.0}{
\begin{tabular}{c|c c c c | c c c c}
\hline
\textbf{System} &  \textbf{COMET-22}  & \textbf{COMETkiwi} & \textbf{ChrF}  & \textbf{BLEU} &
\textbf{COMET-22}  & \textbf{COMETkiwi} & \textbf{ChrF}  & \textbf{BLEU} \\ \hline
& \multicolumn{4}{c}{IS-EN} &   \multicolumn{4}{c}{EN-IS} \\
WMT-Best & \textbf{87.0} &	\textbf{81.4}	& 62.3  &	\textbf{41.7} & \textbf{86.8} &	\textbf{81.8}	& \textbf{59.6} &	\textbf{33.3} \\
MS-Translator &  85.9&	80.3&	\textbf{62.8} &	40.5& 84.3&	80.2&	56.8&	28.7\\\hdashline
GPT  Zeroshot &82.1 &	78.7 &	55.6&	31.9 &76.3 &	74.0 &	43.5 &	15.9\\
GPT 1-Shot RR & 84.1&	80.2&	57.8&	34.7& 77.0 &74.6	&43.7	&15.3 \\
GPT 1-Shot QR & 83.5&	79.7 &	56.7 &	33.3 & 77.4 &	75.1 &	44.5&	16.2 \\
GPT 5-Shot RR &84.4$^*$&	80.4$^*$&	58.1$^*$&	35.0$^*$& 77.9$^*$	&75.2$^*$	&45.1$^*$	&16.8$^*$ \\\
GPT 5-Shot QR & 84.2 &	80.2 &	58.0 &	35.2 & 76.0 &	74.1 &	44.1 &	16.3\\\hline

& \multicolumn{4}{c}{HA-EN} &   \multicolumn{4}{c}{EN-HA} \\
WMT-Best & \textbf{80.0} &	\textbf{74.5} &	\textbf{48.7} &	\textbf{21.0} & \textbf{79.8}&	\textbf{61.5} &	\textbf{51.1} &	\textbf{20.1}\\
MS-Translator &  73.3&	68.5&	43.4&	16.2& 72.5&	57.2&	38.4&	10.3\\\hdashline
GPT  Zeroshot &76.1&	73.1&	45.5&	17.3& 73.3&	58.6&	38.4$^*$&	9.4$^*$\\
GPT 1-Shot RR &75.7 &	72.7 &	45.7 &	17.3 & 74.0 &	59.0 &	38.4$^*$ &	8.8 \\
GPT 1-Shot QR & 78.1 &	74.4 &	47.5$^*$ &	19.1$^*$ & 74.1$^*$ &	59.7$^*$ &	37.8 &	8.9\\
GPT 5-Shot RR & 75.5 &	72.2 &	45.9 &	17.8 & 72.1	 & 57.7	 & 36.0 &	8.0 \\\
GPT 5-Shot QR & 78.2$^*$ &	74.5$^*$ &	47.5$^*$ &	18.9 & 72.6 &	58.5 &	36.9 &	8.5 \\\hline

& \multicolumn{4}{c}{FR-DE} &   \multicolumn{4}{c}{DE-FR} \\
WMT-Best & \textbf{89.5}	& \textbf{80.7} &	\textbf{81.2} &	\textbf{64.8} & \textbf{85.7}	& 79.5 &	\textbf{74.6} &	\textbf{58.4}\\
MS-Translator &85.4&	78.9&	67.5&	45.3 &82.7	&79.0	&65.0&	42.0\\\hdashline
GPT  Zeroshot & 84.6&	77.9&	65.7&	42.5& 78.5	&76.1&	58.9&	35.6\\
GPT 1-Shot RR &86.1 &	79.6 &	65.1 &	41.0 & 83.1 &	80.5 &	60.3	 & 36.9 \\
GPT 1-Shot QR & 86.4 &	80.0 &	67.0 &	43.9 & 83.2 &	80.8	& 61.2	&38.1 \\
GPT 5-Shot RR &  86.6	& 80.0 &	65.2 &	41.6 & 83.6$^*$	& \textbf{80.9}$^*$	& 60.1 &	37.1 \\\
GPT 5-Shot QR & 86.7$^*$ &	80.2$^*$ &	67.7$^*$ &	44.8$^*$  & 83.2 &	80.7	& 62.1$^*$	& 39.3$^*$ \\\hline

\end{tabular}}
\caption{Zero-Shot  and Few-Shots evaluation results with GPT (text-davinci-003) on low resources and non-English centric translation directions from WMT Testsets. The best scores across different systems are marked bold. * denotes the best results among GPT systems.}
\label{tab:result.low-res-xy-wmt}
\vspace{-1.5em}
\end{table*}

We evaluated low-resource and non English-centric languages by conducting experiments with Icelandic and Hausa as two low-resource languages and French and German as direct translation languages. \autoref{tab:result.low-res-xy-wmt} presents the results. The few-shot setup yields modest gains, especially when translating out of English. Similar to the high-resource case, most of the gains were obtained from a single high-quality shot. The systems for both low-resource languages did not surpass the WMT-Best systems. The DE-FR and FR-DE language pairs show remarkable results, as a single-shot setup outperforms the zero-shot setup significantly. This is consistent with the previous finding of translating from English to other languages; a more dense in-context signal is essential for direct translation as well, as it enables the model to generate in the correct language better than the zero-shot behavior. Both direct systems surpasses their commercial counterparts in terms of COMET scores, but they slightly trail behind the  WMT-Best systems on the COMET-22 reference-based metric.

Similar to the high-resource language pairs, both lexical metrics (BLEU and ChrF) showed a significant and consistent degradation. To gain further insights into this, we conducted human evaluation and performed a more in-depth analysis as discussed in~\autoref{sec:humaneval} and~\autoref{sec:analysis}.


\subsection{Document-Level MT with GPT}
\label{sec:document}
This section explores the application of GPT to document-level machine translation. Previous studies on MT with LLMs have mainly concentrated on sentence-level translation, with only a brief mention of document translation for transfer learning by \citet{Zhang-etal-2023-promptingLMs}. Document translation, on the other hand, has received considerable attention for transformer models, with inconclusive findings as to the efficacy of the additional context. \citet{sun-etal-2020-rethinkingdocument} challenge some of the prior studies and demonstrate that a simple modification of the training to vary the document lengths can significantly enhance the performance of a standard transformer architecture for document-to-document translation. We hypothesize that GPT can excel at document-to-document translation, as it is trained on large contexts. Moreover, translating entire documents can reduce the number of API calls and thus improve the computational efficiency and latency. We argue that document translation improvements may need better metrics to capture its potential. 
Hence, in this section, we report results using doc-BLEU \cite{liu2020multilingual} and doc-COMET metrics as described in \autoref{eval}, in addition to our sentence-level metrics.

\paragraph{Evaluating Document-level Translations}
A document-level translation does not necessarily keep sentence-level alignments intact. We try to prompt GPT to keep sentence-level alignment intact by emphasizing sentence separation in the prompt. Our prompt template can be found in~\autoref{fig:prompt_doc} from the appendix. However, we find that we still need to restore sentence-level alignment with source for some of the documents in the test set. In all cases, we find two types of mismatch that we need to restore. First, sentences that were written in the source over two lines and their translation is one line. In that case, we insert a new line break to match the position of the new line break in the source sentences. Second, sentences that are skipped. In that case, we replace empty lines at the end of the documents with empty lines in place of the skipped sentences. 

We need to restore sentence-level alignment to calculate metrics that were mainly developed for sentence-level evaluation such as the COMET-22 and COMETkiwi  models that we are using. We also follow \cite{liu2020multilingual} and report SacreBLEU calculated on the document level. For neural network based metrics, we extend the COMETkiwi  model for document level evaluation as described in~\autoref{eval}. 


\paragraph{Experiment 1:} We conduct a series of experiments on zero-shot MT using News Commentary dataset, varying the window length from 1 (sentence-level) to 32 in powers of two\footnote{The context is limited to the document if the window length exceeds the document length.}. \autoref{tab:result.doc-wmt-batch} 
shows that increasing the window size leads to improvements across all metrics. However, the gains in lexical metrics (BLEU and ChrF) are larger than in neural metrics (COMET-22 and COMETkiwi). The document-based metric (doc-BLEU and doc-COMETkiwi) also exhibit similar improvements to the sentence-based metric. Remarkably, as the window size grows, the performance surpasses MS-Translator model and approaches the  WMT-Best systems. This is consistent with the findings of~\citet{sun-etal-2020-rethinkingdocument} for conventional MT models. The table also displays the total number of requests for each window size. We observe that the number of requests decreases dramatically as the window size increases, while the performance either improves significantly or remains relatively stable depending on the metric used. Therefore, this document-level setup  achieves high efficiency without compromising the quality.

\begin{table*}[!htbp]
\centering
\small
\vspace{-0.8em}
\scalebox{0.9}{
\begin{tabular}{c|c c c c c c c }
\hline
\textbf{System} &  \textbf{COMET-22}  & \textbf{COMETkiwi} & \textbf{Doc-COMETkiwi} & \textbf{ChrF}  & \textbf{BLEU}  & \textbf{Doc-BLEU}  & \textbf{GPT Requests}   \\ \hline
& \multicolumn{7}{c}{DE-EN} \\

WMT-Best & 85.0 &	\textbf{81.4} & 79.9 &	\textbf{58.5} &	33.4 & \textbf{35.2} & -- \\
MS-Translator &84.7&	81.0 & 79.5 &	\textbf{58.5}&	\textbf{33.5} & \textbf{35.2} & -- \\
GPT Sent ZS & 84.8	&81.2 	& 79.5 & 56.8 &	30.9 & 32.3 & 1984 \\ \hdashline

GPT  Doc ZS w=2 & 85.1 &  \textbf{81.4}$^*$ & 80.0 & 57.8 & 32.6 & 34.4 & 1055 \\
GPT  Doc ZS w=4  &  \textbf{85.2}$^*$ &  81.3 & \textbf{80.2}$^*$ & 57.9 & 32.8 & 34.5 & 607 \\
GPT Doc ZS w=8 &  85.1 &  81.2 & 80.2 & 57.9 & 33.0 & 34.7 & 401 \\
GPT  Doc ZS w=16 &  85.2 &  81.2 & 80.2 & 58.0$^*$ & 33.1$^*$ & 34.8$^*$ & 310 \\
GPT   Doc ZS w=32 &  85.1 &  81.2 & 80.2 & 57.9 & 33.1 & 34.8 & 274 \\ \hline

& \multicolumn{7}{c}{EN-DE} \\
WMT-Best & \textbf{87.2} & \textbf{83.6} &	\textbf{83.1} & \textbf{64.6} & \textbf{38.4} & \textbf{40} & -- \\
MS-Translator & 86.8  & 83.4  & 83	& 64.2 & 37.3 & 38.8 & -- \\
GPT Sent ZS & 85.6  & 82.8 & 82.2 & 60.2  & 31.8 & 33.1 & 2037 \\ \hdashline

GPT  Doc ZS w=2 & 86.1  & 82.7 &  82.4 & 60.9 & 32.8 & 34.4 & 1058 \\
GPT  Doc ZS w=4 & 86.3 & 82.6  &  82.6 & 61.3 & 33.6 & 35.2 & 579 \\
GPT  Doc ZS w=8 & 86.4 & 82.6 & 82.6 & 60.9 & 33.4 & 35.2 & 349 \\
GPT  Doc ZS w=16 & 86.5$^*$  & 82.6$^*$  & 82.6$^*$  & 61.3$^*$ & 34.2$^*$ & 36.1$^*$ & 235 \\
GPT  Doc ZS w=32 & 86.4  & 82.6  &  82.7 & 61.3 & 34.1 & 36.1 & 187 \\\hline

\end{tabular}}
\caption{Evaluation results of document-level translation with GPT on DE<>DE WMT22 testset. The table shows the effect of increasing context length $w$ in document to document translation with a zero-shot setting.}
\label{tab:result.doc-wmt-batch}
\vspace{-1.5em}
\end{table*}

\paragraph{Experiment 2:} The second set of experiments investigates few-shot translation for the document setting. Following the sentence experiments, we focus on using 5-shots. We also use the News Commentary dataset, which has document-level annotations. \autoref{tab:result.doc-wmt-shots} summarizes the results. The first two rows show the best WMT22 and the MS-Translator results for reference.

The following rows are named GPT-XX-YY where XX stands for the scope of translation (sentence or document) and YY stands for the source of the 5 shots (QR,DR,DF or DH as explained below).
Rows  GPT-Sent-QR and GPT-Sent-DR show results for sentence-level translation. The former uses the same quality-based shots from \autoref{tab:result.few-full-wmt}, while the latter uses 5 randomly selected shots from the document set, excluding the test data. We perform document translation (referred to as Doc in the table) for a window of 10 sentences in the following rows. Rows named GPT-Doc-QR and GPT-Doc-DR use the same shots as the sentence case. For row GPT-Doc-DF, we select a random document from the document data pool  and use the first 5 sentences of the document as shots (i.e. document first DF). For row GPT-Doc-DH, we store GPT outputs in history for use in shots. We translate the first document 0-shot, and the subsequent documents 5-shots. For shot selection, we pick a random document from the previously translated documents and use the first 5 sentences as shots (i.e. document history DH). The results indicate that document translation outperforms sentence translation across metrics. However, while few-shots yield some consistent gains for sentence translation, this is not the case for document translation. This may be explained by the fact that document translation provides enough context, making few-shots redundant.
It can be also observed from the table that the Doc-COMETKiwi shows more gain than the sentence level metrics but this might need more indepth analysis to verify.

\begin{table*}[!htbp]
\centering
\vspace{0.8em}
\scalebox{0.9}{
\begin{tabular}{c|c c c c c c}
\hline
\textbf{System} &  \textbf{COMET-22}  & \textbf{COMETkiwi} & \textbf{Doc-COMETkiwi} & \textbf{ChrF}  & \textbf{BLEU} & \textbf{Doc-BLEU} \\ \hline
& \multicolumn{6}{c}{DE-EN} \\  

WMT-Best                & 85.0 & 81.4  & 79.9  & \textbf{58.5}  & 33.4  & \textbf{35.2} \\
MS-Translator             & 84.7 & 81.0  & 79.5  & \textbf{58.5}  & \textbf{33.5}  & \textbf{35.2} \\\hdashline

GPT-Sent-QR        & \textbf{85.4}$^*$  & \textbf{81.5}$^*$  & 80.2  & 57.7  & 32.4  & 34.0 \\
GPT-Sent-DR    & 84.9 & 81.4  & 80.0  & 55.3  & 29.6  &   31.3 \\\hdashline

GPT-Doc-QR            &  85.2 & 81.2  & 80.2  & 58.1  & 33.3  & 35.0  \\
GPT-Doc-DR            &  85.2 & 81.3  & \textbf{80.5}$^*$  & 57.6  & 32.7  &  34.3  \\
GPT-Doc-DF  &  85.1 & 81.3  & 80.4  & 57.3  & 32.6  &  34.1 \\
GPT-Doc-DH       &  85.3 & 81.2  & 80.3  & 58.2$^*$  & \textbf{33.5}$^*$  &  35.1$^*$  \\\hline

&\multicolumn{6}{c}{EN-DE} \\

WMT-Best                & \textbf{87.2} & \textbf{83.6}  &  \textbf{83.1}  &  \textbf{64.6}  &  \textbf{38.4}  &  \textbf{40.0} \\
MS-Translator             & 86.8 & 83.4  & 83.0  & 64.2  & 37.3  &   38.2 \\\hdashline

GPT-Sent-QR        & 86.4 & 83.1  & 82.7  & 61.3  & 33.2  & 34.8 \\
GPT-Sent-DR    & 86.7 & 83.5$^*$  & 83.0$^*$  & 61.2  & 32.8  &   34.3 \\\hdashline

GPT-Doc-QR             & 86.6 & 83.0  & 82.7  & 61.6  & 34.0  &  35.7 \\
GPT-Doc-DR             & 86.9 & 83.0  & 82.9  & 61.9  & 34.4  &  36.1  \\
GPT-Doc-DF   &  87.0$^*$ & 83.1  & 82.9  & 62.0$^*$  & 34.5$^*$  & 36.2$^*$  \\
GPT-Doc-DH        & 86.6 & 82.9  & 82.8  & 61.7  & 33.9  &  35.7  \\\hline

\end{tabular}}
\caption{Effect of shot selection for document-level translation on WMT22 DE<>EN testset.}
\label{tab:result.doc-wmt-shots}
\vspace{-1.5em}
\end{table*}

\subsection{Robustness Toward Domain Shift}
\label{sec:domains}

\begin{table*}[!htbp]
\centering
\vspace{-0.8em}
\scalebox{0.9}{
\begin{tabular}{c|c c c c | c c c c}
\hline
\textbf{System} &  \textbf{COMET-22}  & \textbf{COMETkiwi} & \textbf{ChrF}  & \textbf{BLEU} &
\textbf{COMET-22}  & \textbf{COMETkiwi} & \textbf{ChrF}  & \textbf{BLEU} \\ \hline
&  \multicolumn{8}{c}{\emph{Combined All Domains}} \\
& \multicolumn{4}{c}{DE-EN} &   \multicolumn{4}{c}{EN-DE} \\ 

WMT-Best & 85.0 &	81.4 &	58.5 &	33.4 & \textbf{87.2} & \textbf{83.6} &	\textbf{64.6} &	\textbf{38.4}\\
MS-Translator &84.7&	81.0&	\textbf{58.5}&	\textbf{33.5}& 86.8&	83.4&	64.2&	37.3 \\
GPT  & \textbf{85.4}	& \textbf{81.5} &	57.7&	32.4 &86.4&	83.1&	61.3&	33.2 \\\hdashline

& \multicolumn{4}{c}{ZH-EN} &   \multicolumn{4}{c}{EN-ZH} \\
WMT-Best & 81.0	& 77.7 & 	\textbf{61.1} &	\textbf{33.5} & 86.7	& 82.0	 & \textbf{41.1} &	\textbf{44.8}\\
MS-Translator &80.4& 77.6	& 57.7 & 	27.9 & 86.1 & 	81.4	& 43.1 & 	48.1\\ 
GPT   & \textbf{81.1} &	\textbf{78.7} &	54.7 &	23.8 & \textbf{87.0}	& \textbf{82.2}	& 39.8	 & 43.7 \\\hline

&  \multicolumn{8}{c}{\emph{Conversational  Domain}} \\
& \multicolumn{4}{c}{DE-EN} &   \multicolumn{4}{c}{EN-DE} \\ 

WMT-Best & 85.7 &	81.4 &	54.9 &	35.1  & \textbf{89.1} & 83.3 &	\textbf{67.7} &	\textbf{42.8}\\
MS-Translator & 85.1 & 81.0 & 	\textbf{55.2} &	35.3 & 88.8&	83.1&	67.3 &	40.7 \\
GPT   & \textbf{86.1}&	\textbf{81.5} &	55.0 &	\textbf{35.6}  & 88.5 &	\textbf{83.4} &	62.9 & 35.7\\\hdashline
		
& \multicolumn{4}{c}{ZH-EN} &   \multicolumn{4}{c}{EN-ZH} \\
WMT-Best & 81.6	& 77.7 & 	\textbf{48.8}  &	25.6  & 86.9	& 81.3	 & 36.7 &		37.6\\
MS-Translator &80.6 & 77.4	& 46.7 & 	22.9 & 87.6 & 	81.3	& \textbf{42} & 	\textbf{44.0} \\ 
GPT & \textbf{82.0} &	\textbf{78.1} &	48.0 &	\textbf{26.3}  & \textbf{88.9}	& \textbf{82.4}	&40.5	 & 41.9 \\\hline

&  \multicolumn{8}{c}{\emph{News  Domain}} \\
& \multicolumn{4}{c}{DE-EN} &   \multicolumn{4}{c}{EN-DE} \\ 

WMT-Best & 84.9  &	82.0  &	58.8 &	31.4 & \textbf{87.0} &	\textbf{84.5} & \textbf{65.6}  & \textbf{37.8}\\
MS-Translator &84.7&	81.9 &	\textbf{59.0} &	\textbf{31.6} & 87.0	& 84.3  & 65.3  & 36.8 \\
GPT   & \textbf{85.3} &	\textbf{82.3} &	58.7 &	31.4 & 85.9 &	83.7 & 62.4 & 31.7\\\hdashline

& \multicolumn{4}{c}{ZH-EN} &   \multicolumn{4}{c}{EN-ZH} \\
WMT-Best & 	\textbf{82.1} &	79.6 & \textbf{62.2} & \textbf{31.3} & \textbf{87.6} & \textbf{83.1} & 45.6 & 51.7 \\
MS-Translator & 81.8 &	80.0 & 59.7 & 28.2 & 87 & 82.3 & \textbf{48.2} & \textbf{53.8} \\ 
GPT  & 81.7 & \textbf{80.2} & 56.9 & 23.3 & 87.2 & 82.5 & 42.2 & 48.7 \\\hline

& \multicolumn{8}{c}{\emph{e-Commerce  Domain}} \\
& \multicolumn{4}{c}{DE-EN} &   \multicolumn{4}{c}{EN-DE} \\ 
WMT-Best & \textbf{85.6}	& \textbf{81.2}	 & \textbf{61.3} & \textbf{35.3}  &		\textbf{88.7}	& \textbf{84.1} & \textbf{65.3} &  \textbf{38.4} \\
MS-Translator &85.3 &	80.9 &	\textbf{61.3} &  35.2 & 	88.2 &	83.8 & 64.9 & 38.1 \\
GPT   & \textbf{85.6}	& \textbf{81.2} & 60.5 & 34	& 87.6 &	83.6 & 62 & 34 \\\hdashline

& \multicolumn{4}{c}{ZH-EN} &   \multicolumn{4}{c}{EN-ZH} \\
WMT-Best & 77.6 & 75.1 & \textbf{52.2} & \textbf{22.2}&  \textbf{88.0}	& \textbf{83.0} & 40.8 & 43.5\\
MS-Translator &77.8 & 	75.0 &  51.5 & 20.3  &			87.8 & 	82.6 &  \textbf{41.9} & \textbf{46.6} \\ 
GPT  & \textbf{79.0} &	\textbf{76.7} 	& 49.4 & 18.7  & 	87.9	& 82.8  &40.3  & 42.9 \\\hline

&  \multicolumn{8}{c}{\emph{Social Domain}} \\
& \multicolumn{4}{c}{DE-EN} &   \multicolumn{4}{c}{EN-DE} \\ 

WMT-Best & 84.1 &	80.9  & \textbf{56.5} &  \textbf{32.6}	&		\textbf{84.0} &	\textbf{82.3}  & \textbf{59.8} & \textbf{35.8}  \\
MS-Translator & 83.7 & 	80.5	 & 56.2 & 32.5 	& 83.2 & 82.2 & 59.1 & 34.6	 \\
GPT  & \textbf{84.5} &	\textbf{81.1} & 	54.7 & 29.9	& 	83.6 &	81.7 & 57.7 & 32.5 \\\hdashline

& \multicolumn{4}{c}{ZH-EN} &   \multicolumn{4}{c}{EN-ZH} \\
WMT-Best & \textbf{83.0} &	78.2 &	\textbf{69.4} &  \textbf{46.8} &	\textbf{84.2} &	 80.7 & \textbf{36.4} &  40.2\\
MS-Translator &81.4	& 78.0	& 62.3  & 34.6 & 		82.0 &	79.4  & 36.1 & \textbf{40.9}\\ 
GPT  & 81.9	 & \textbf{79.7} & 57.4 & 28 		&	84.0 &	\textbf{81.0} & 34.2 & 37.7  \\\hline

\end{tabular}}
\caption{Evaluation results of DE<>EN and ZH<>EN translations across four domains.}
\label{tab:result.domains-wmt}
\vspace{-1.5em}
\end{table*}


We use the WMT datasets to examine how domain shift affects the performance of GPT models on German and Chinese, both from and to English. The WMT22 datasets span four domains: Conversational, News, e-Commerce and Social.  \autoref{tab:result.domains-wmt} shows the scores of the four domains on the WMT testsets.

GPT achieves remarkable improvements on the conversational domain for DE-EN, ZH-EN and EN-ZH, as evidenced by both COMET and lexical scores (BLEU and ChrF). This contrasts with previous observations where lexical scores consistently deteriorate with GPT models.

GPT performs comparably to the other systems on the news domain according to COMET scores for all directions. It surpasses both other systems on DE-EN, while slightly trailing behind on EN-DE. For ZH-EN and EN-ZH, GPT exceeds MS-Translator, but falls slightly short of WMT-Best system.
However, GPT scores significantly lower in terms of BLEU metric for both ZH-EN and EN-DE.

GPT clearly outperforms both systems on ZH-EN and matches WMT-Best on DE-EN for the e-commerce domain. It slightly lags behind on other directions. In this domain, we observe consistent lower scores in BLEU metric for all directions even for ZH-EN which outperforms significantly on both COMET metrics.

GPT outperforms both systems on DE-EN for the social domain. However, on ZH-EN and EN-ZH, GPT only surpasses them on COMETkiwi, while showing lower BLEU score for all directions with a significant difference in ZH-EN which exhibits substantial gains on COMETkiwi.

The results demonstrate GPT’s robust translation capabilities across different domains and languages. It performs well on DE-EN, ZH-EN and EN-ZH for all domains. However, we observe a discrepancy on lexical metrics for ZH-EN and DE-EN on News and Social domains even with GPT’s high performance on those languages. We conduct a further examination of the ZH-EN results to gain more insights. We find that the news from Chinese outlets follows a more templatic style especially in the prefix of the news. For NMT systems that are trained heavily on similar data, it is easier to reproduce the same patterns, e.g., WMT-Best scores  31.3 BLEU. For more general commercial scale systems that are trained on much larger and diverse data, it is harder to produce the same exact patterns, as such,  MS-Translator scores 28.2 BLEU. For GPT, which is mostly trained on English, it is harder to get the lexical matches and it mostly produces the English news style, scoring 23.3 BLEU.  However, COMET-22,  using the same reference, seems to provide a more robust signal with all three systems being almost on the same level. This confirms the robustness of neural metrics across domains \cite{metrics22} and GPT’s ability to handle diverse domains while being more robust towards \textit{parallel data biases}, which we will explore further in \autoref{sec:analysis}. Therein, we show that in general, GPT performs better in cases where the input \textit{resonates} with the noisy parts of parallel data.



\begin{figure*}[h!]
  \centering

         \includegraphics[width=0.8\textwidth]{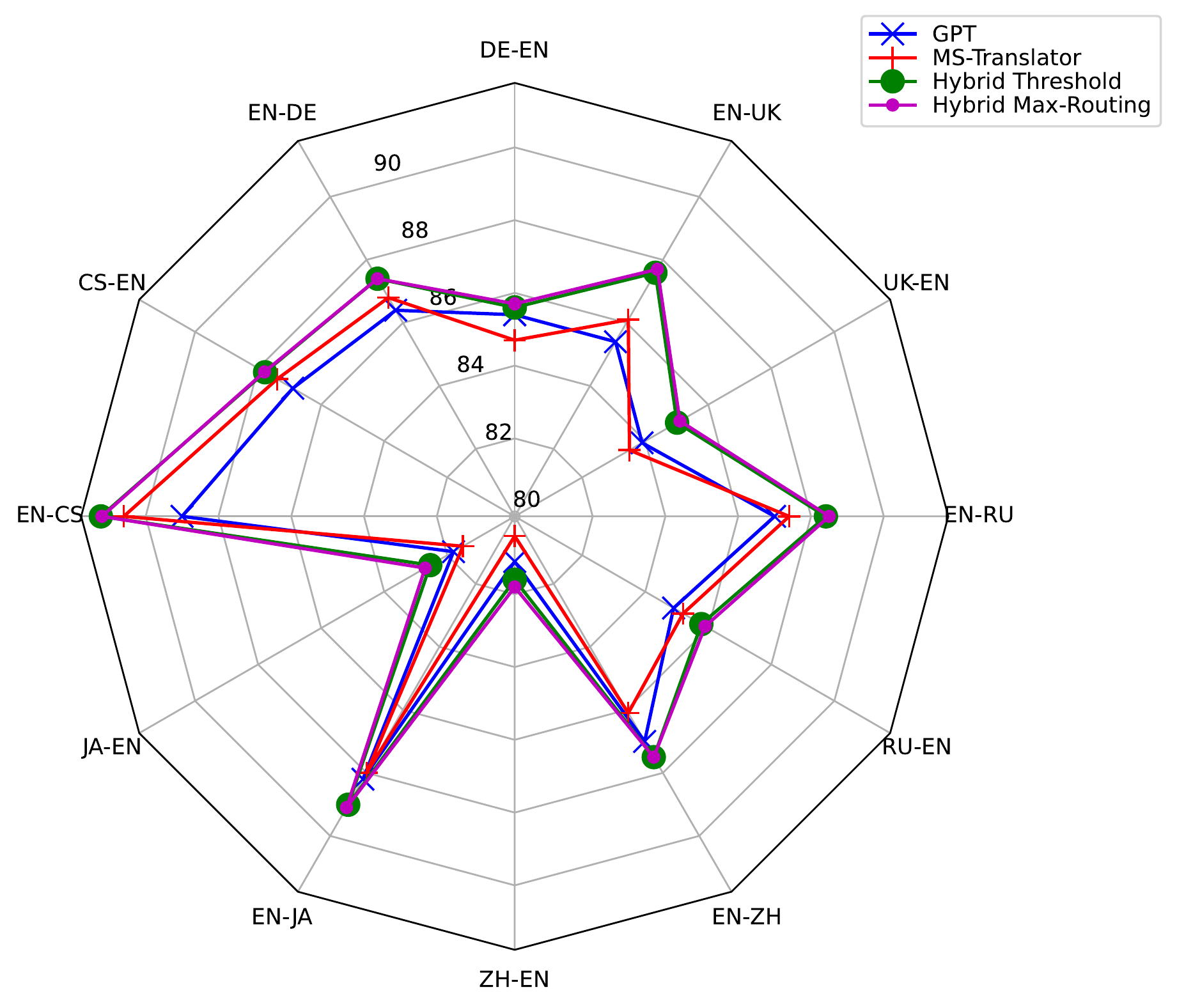}
        \vskip -0.1in
        \caption{\centering  Comparing COMET-22 scores of hybrid MS-Translator and GPT systems with GPT and MS-Translator systems.}
         \label{fig:hybrid-gpt}
  
\end{figure*}

\begin{figure*}[h!]
    \centering
    \begin{subfigure}[b]{0.48\textwidth}
        \centering
        \includegraphics[width=1.0\textwidth]{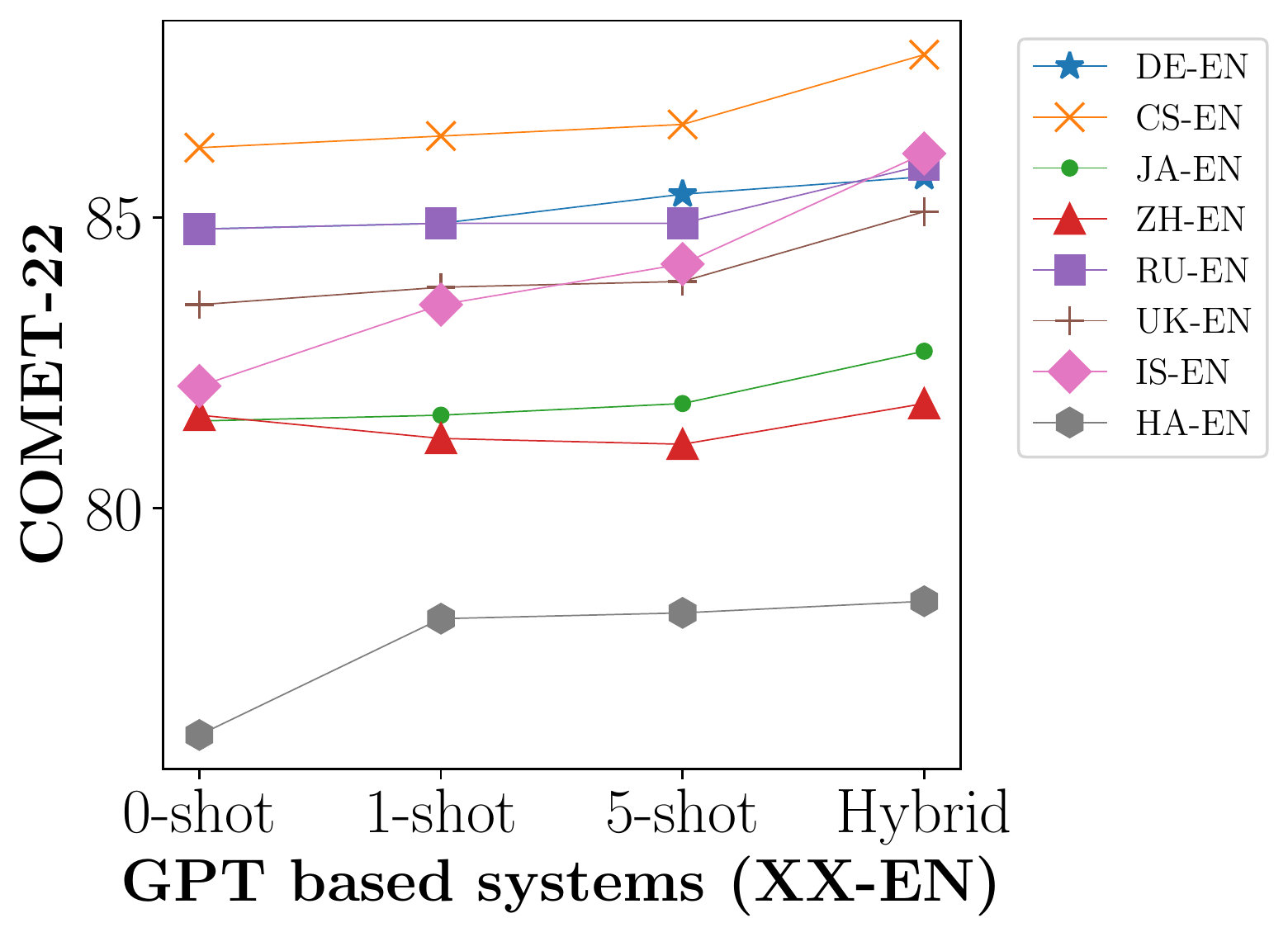}
        \vskip -0.05in
        \label{fig:xx-en-comet22}
        \end{subfigure}
    \begin{subfigure}[b]{0.48\textwidth}
        \centering
        \includegraphics[width=1.0\textwidth]{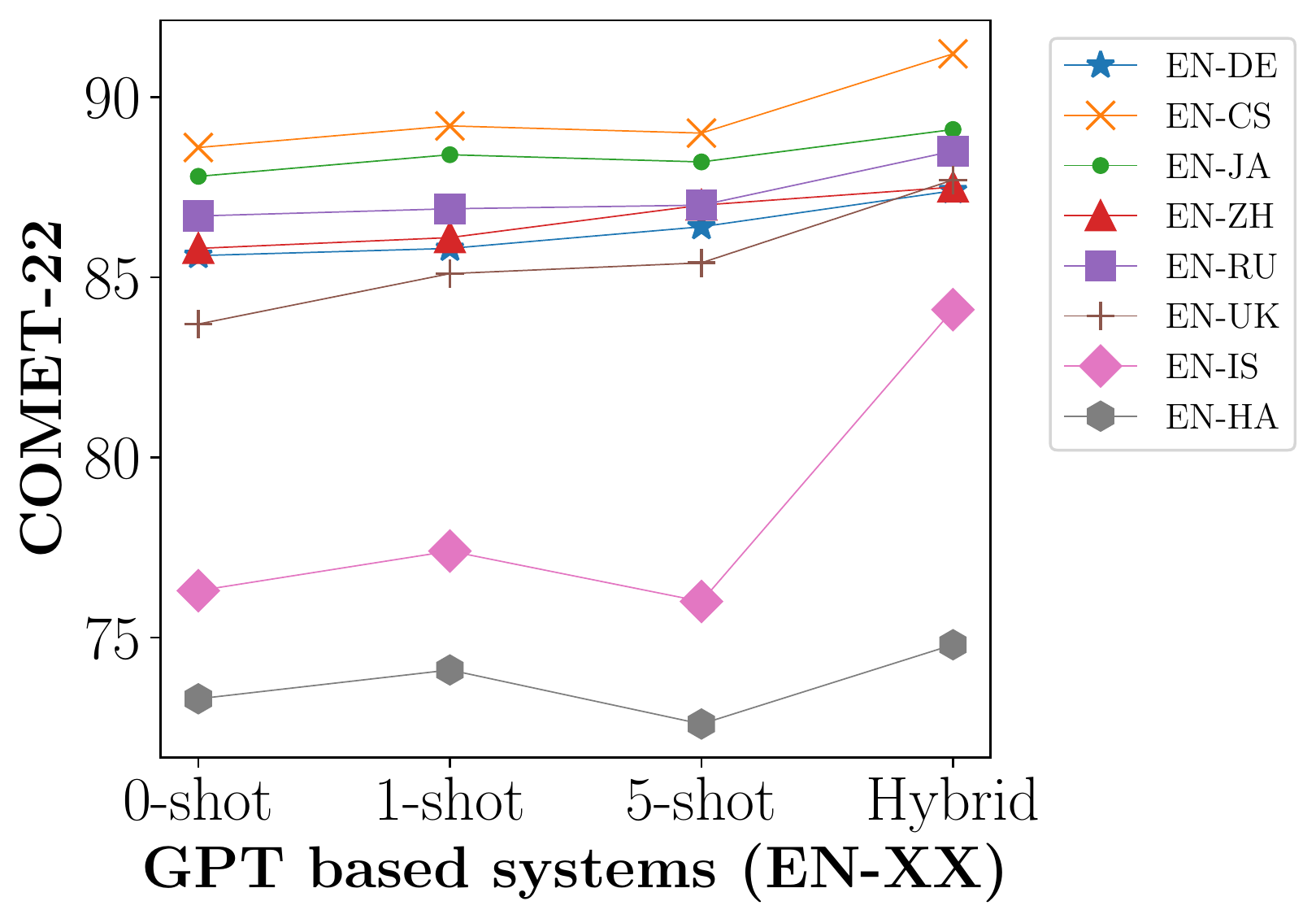}
        \vskip -0.05in
        \label{fig:en-xx-comet22}
    \end{subfigure}
    \begin{subfigure}[b]{0.48\textwidth}
        \centering
        \includegraphics[width=1.0\textwidth]{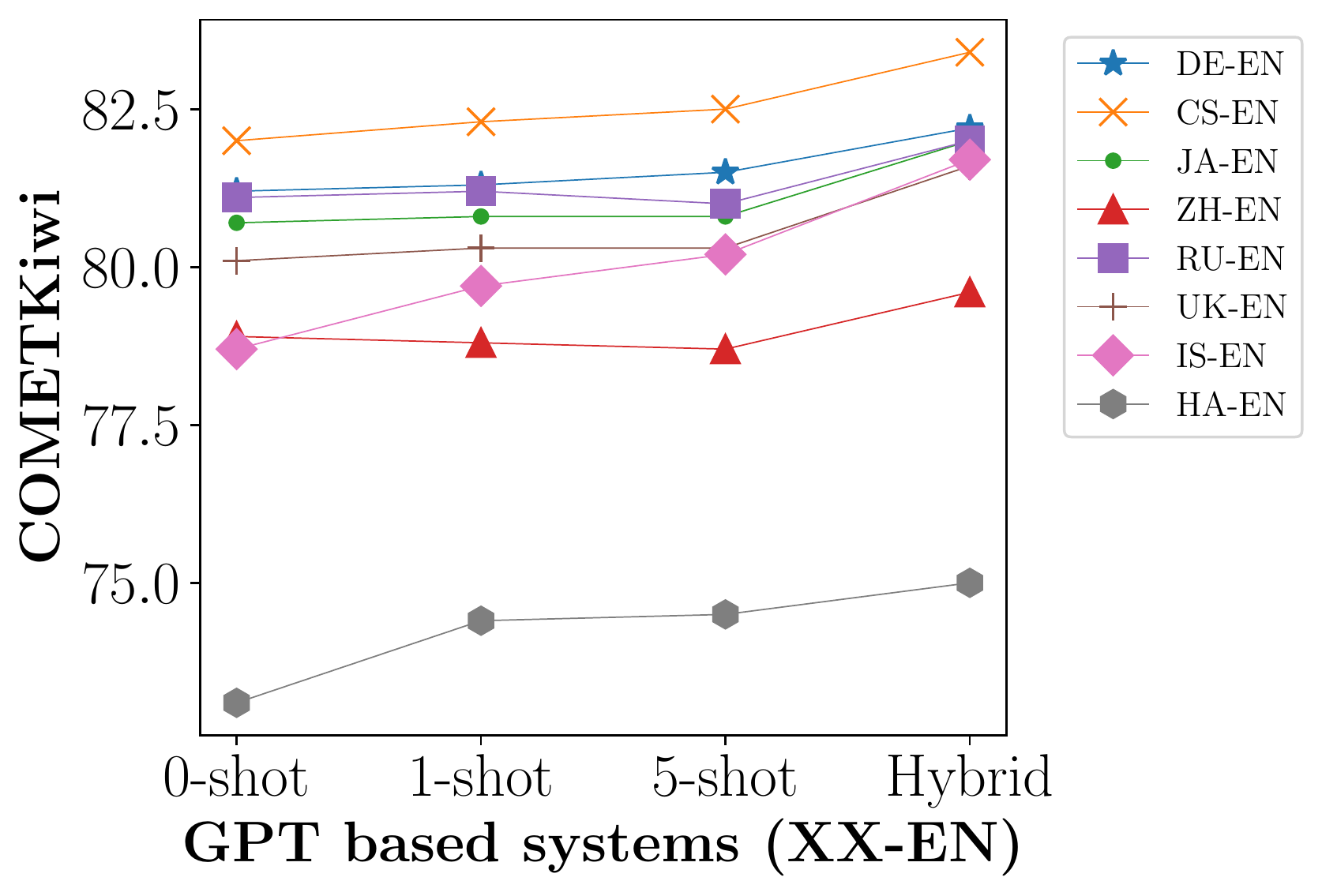}
        \vskip -0.1in
        \label{fig:xx-en-COMETqe}
        \end{subfigure}
    \begin{subfigure}[b]{0.48\textwidth}
        \centering
        \includegraphics[width=1.0\textwidth]{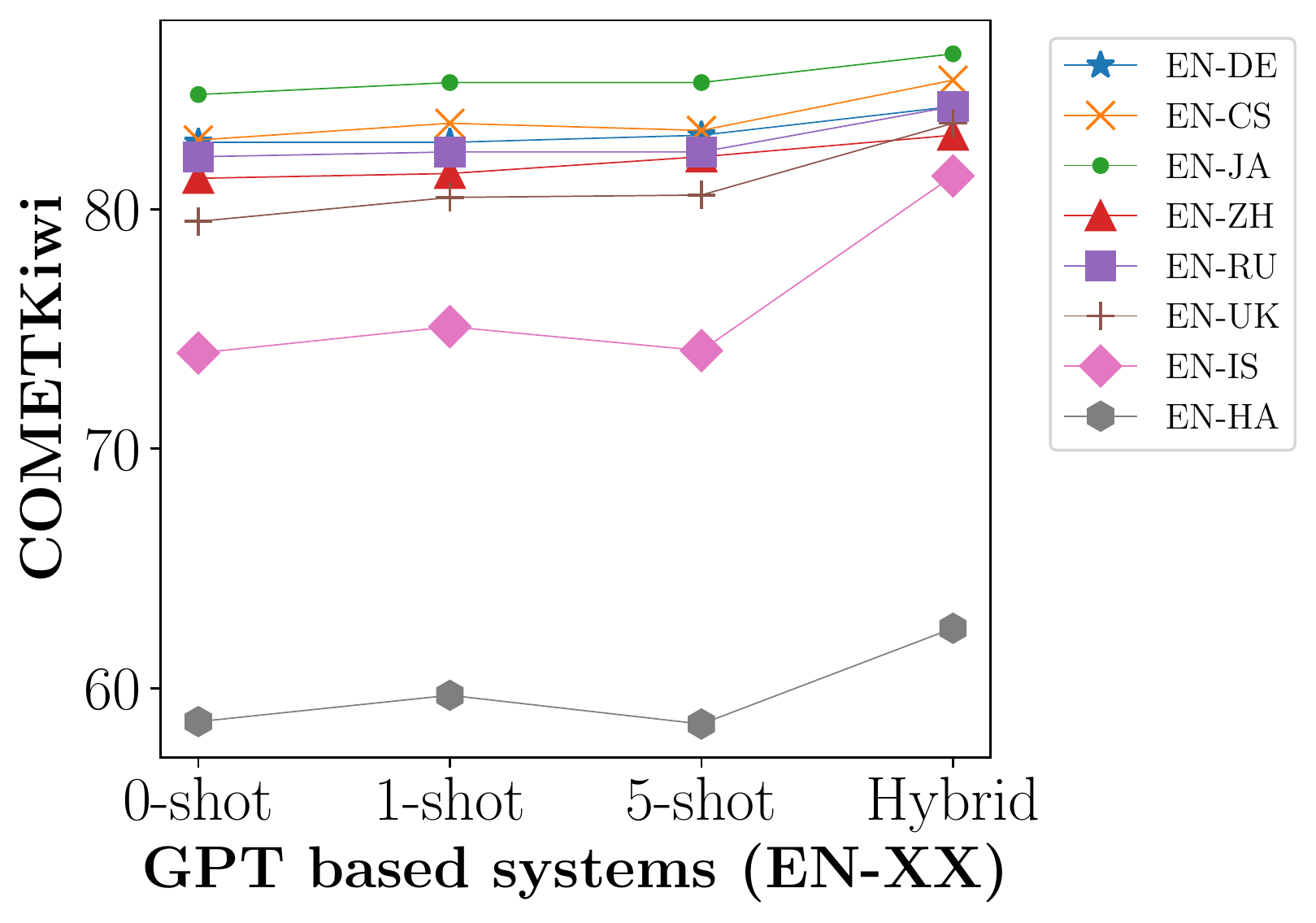}
        \vskip -0.1in
        \label{fig:en-xx-COMETqe}
    \end{subfigure}
    \caption{COMET-22 and COMETKiwi scores for GPT based systems with different approaches.}
    \label{fig:enxxCOMET22}
\end{figure*}

\begin{figure*}[h]
    \centering
    \begin{subfigure}[b]{0.45\textwidth}
        \centering
        \includegraphics[width=1.0\textwidth]{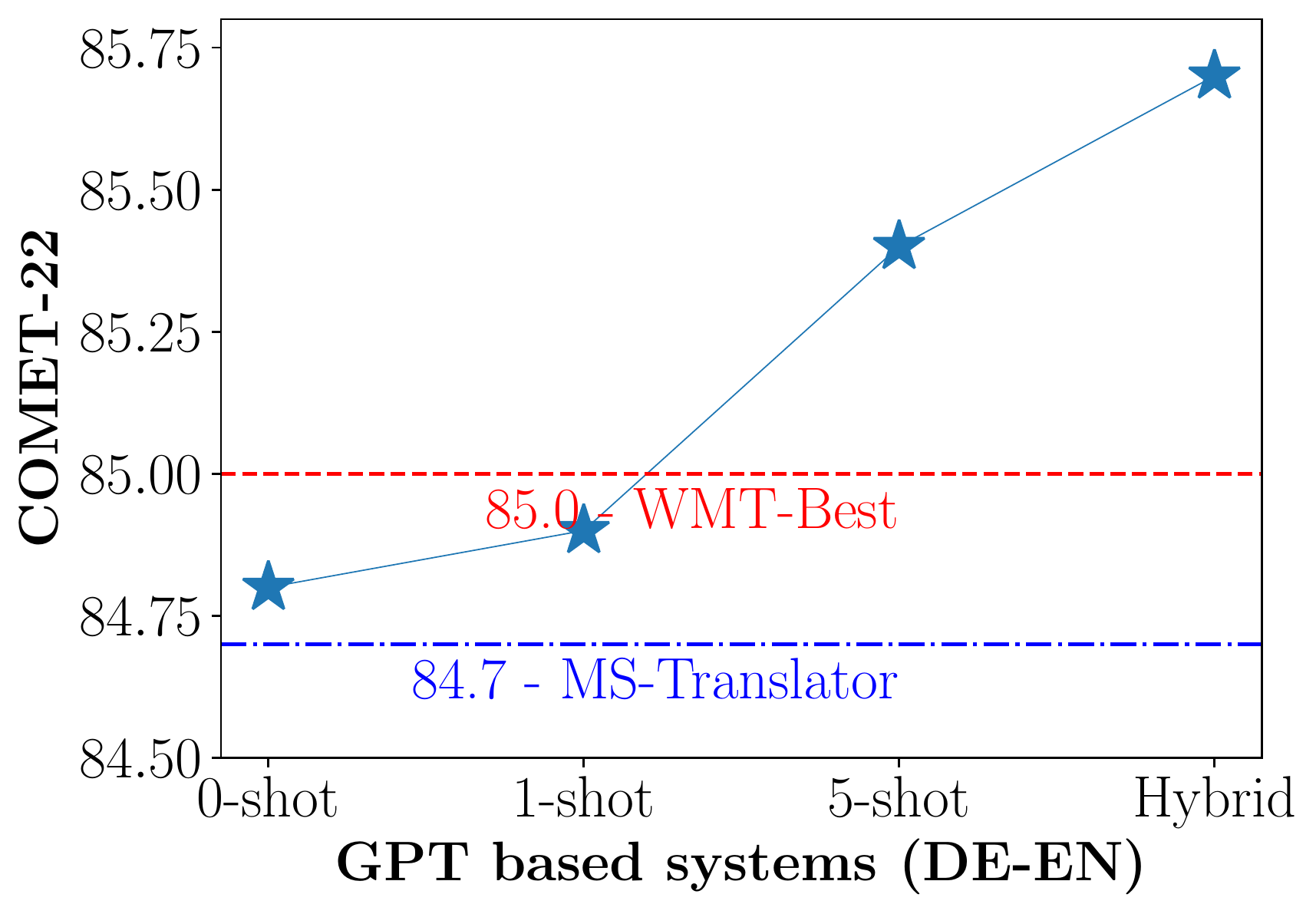}
        \vskip -0.05in
        \label{fig:de-en-COMET22}
        \end{subfigure}
    \begin{subfigure}[b]{0.45\textwidth}
        \centering
        \includegraphics[width=1.0\textwidth]{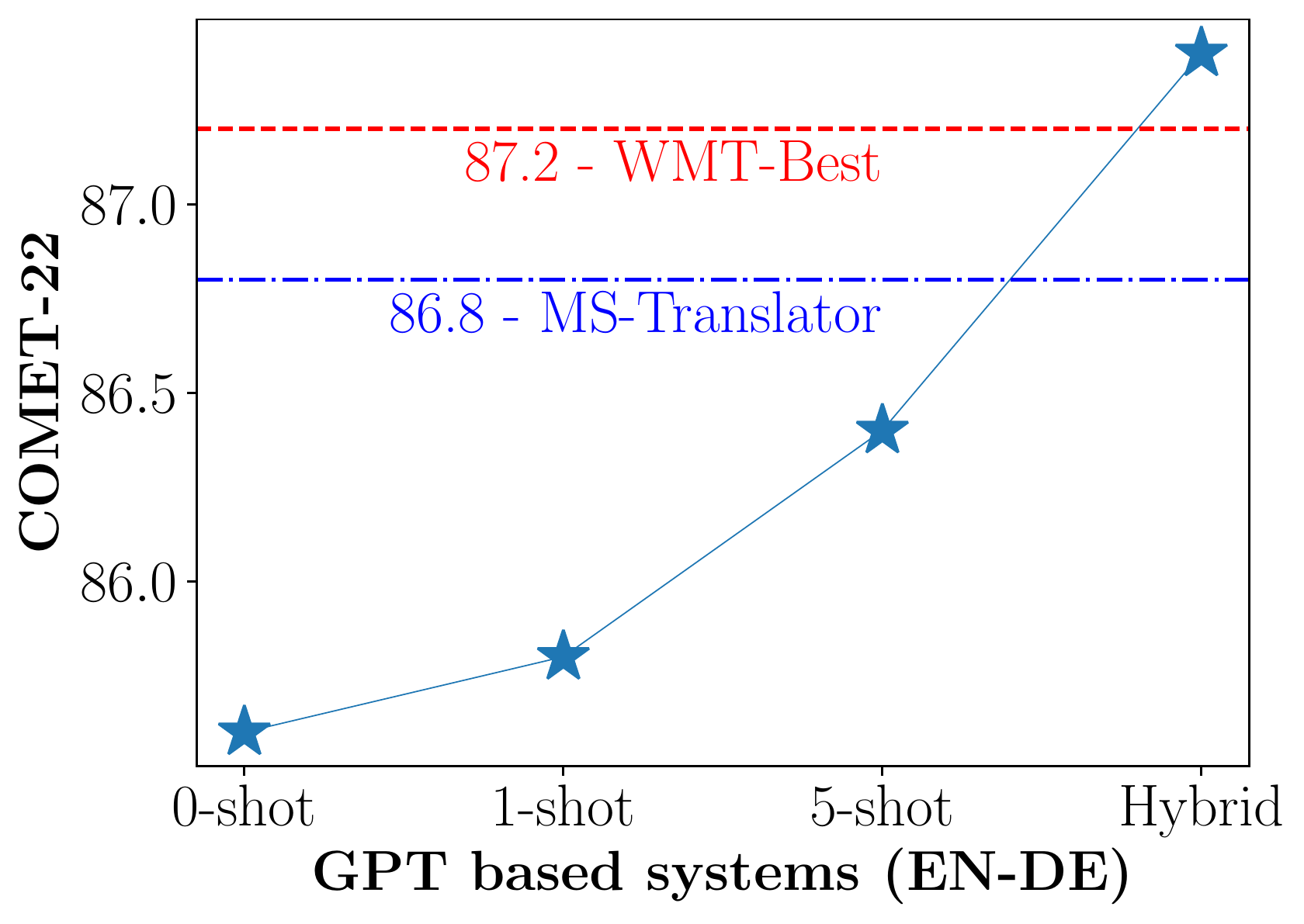}
        \vskip -0.05in
        \label{fig:en-de-COMET22}
    \end{subfigure}
    \begin{subfigure}[b]{0.45\textwidth}
        \centering
        \includegraphics[width=1.0\textwidth]{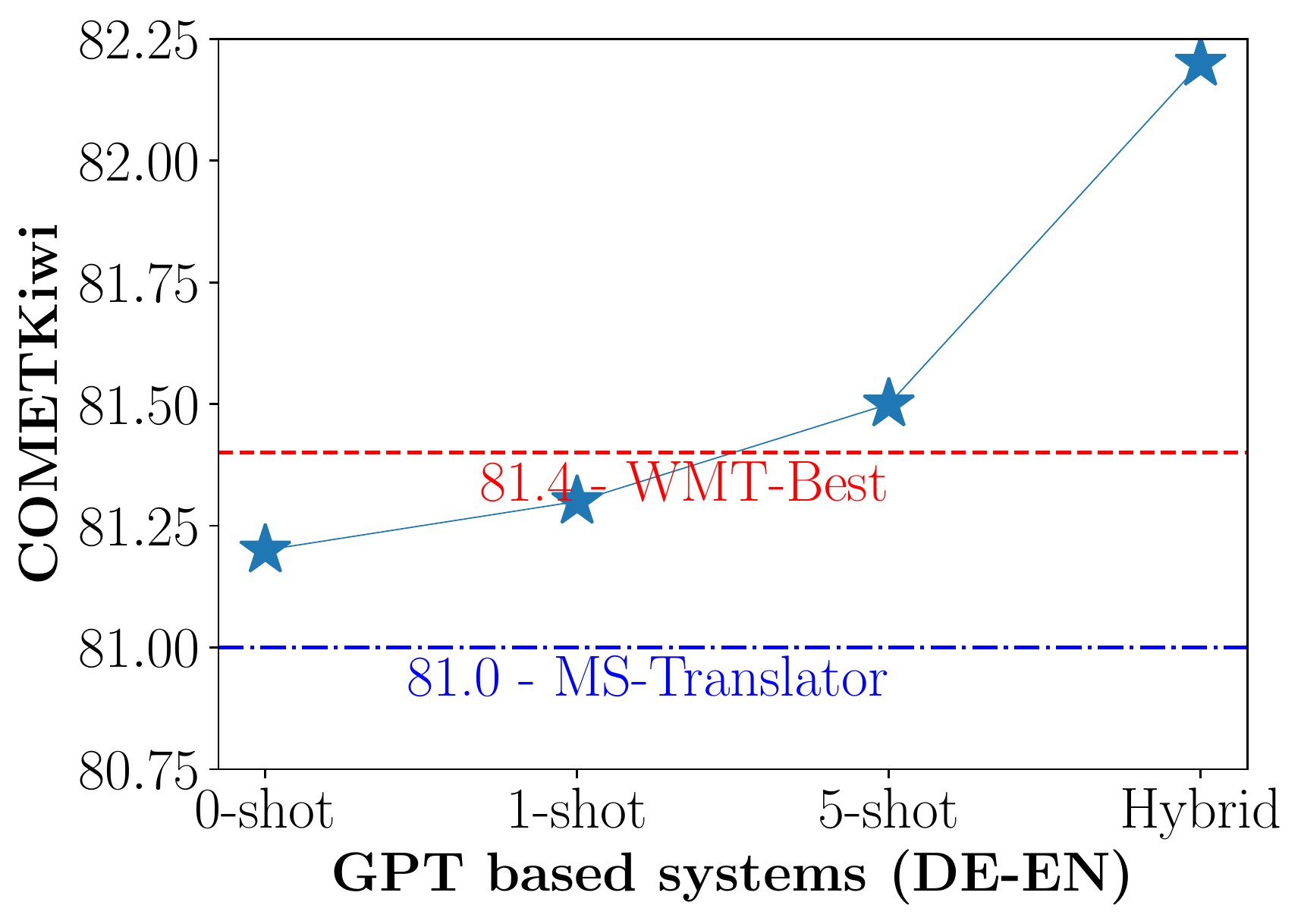}
        \vskip -0.1in
        \label{fig:de-en-COMETqe}
        \end{subfigure}
    \begin{subfigure}[b]{0.45\textwidth}
        \centering
        \includegraphics[width=1.0\textwidth]{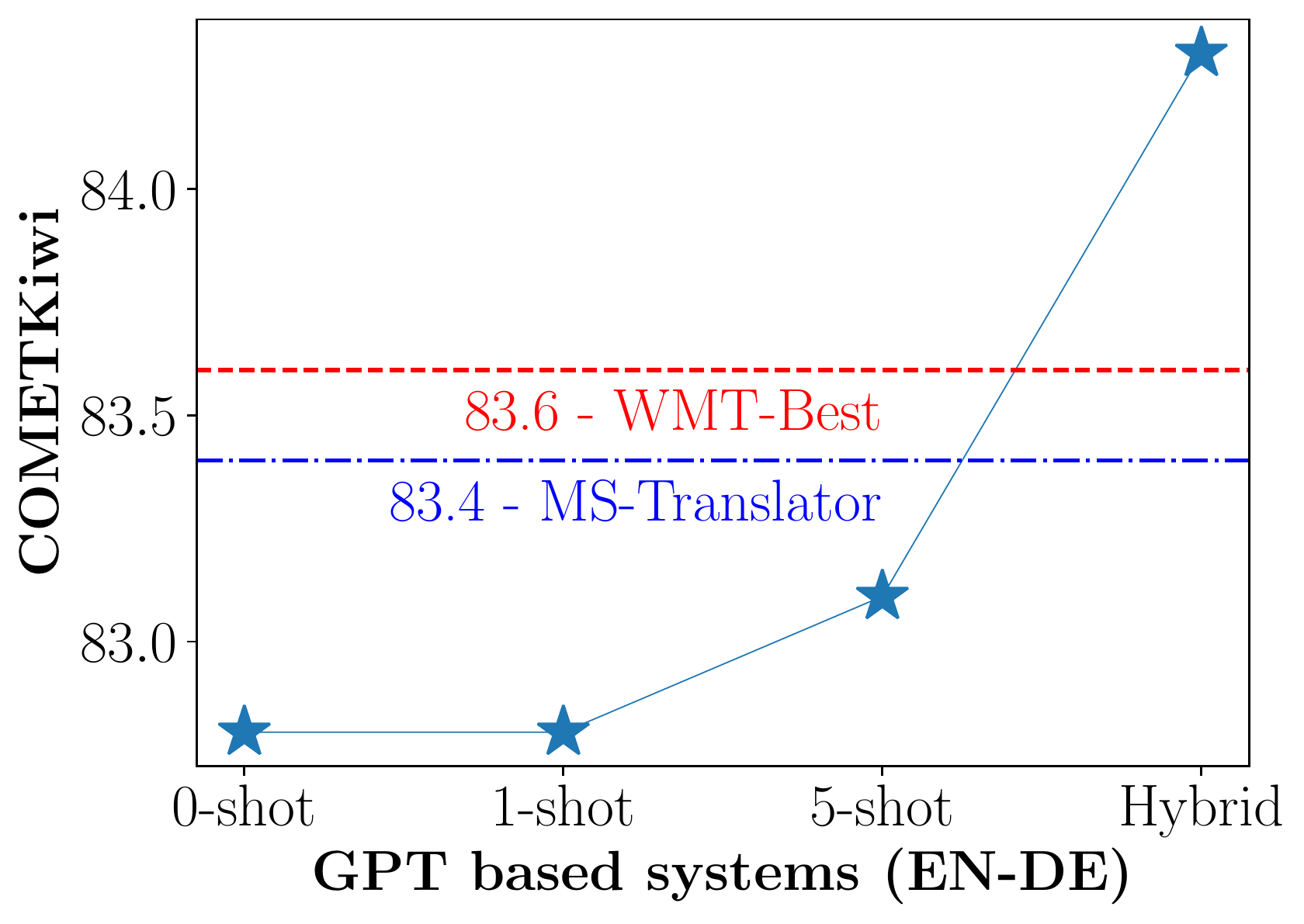}
        \vskip -0.1in
        \label{fig:en-de-COMETqe}
    \end{subfigure}
    \caption{COMET-22 and COMETKiwi scores for GPT based systems compared to WMT-Best and MS-Translator systems to translate between English (EN) and German (DE).}
    \label{fig:endeCOMET22}
\end{figure*}

\subsection{Hybrid GPT and NMT Translation}
\label{sec:hybrid}

To explore the possibility of leveraging the strong performance of GPT on various languages, we propose and evaluate several hybrid approaches that combine the strengths of NMT and GPT systems. The basic idea is to use Microsoft Translator (MS-Translator) system as the primary translation system, and then use GPT as a fallback system when the quality of MS-Translator is unsatisfactory.

We use COMETkiwi as the quality estimation model and COMET-22 as the performance evaluation metric. We first establish an upper bound by selecting the best translation from either systems according to COMETkiwi, which we call the ``Max-Routing'' approach. Then we experiment with a more practical approach where we use GPT only when the COMETkiwi score of MS-Translator falls below a predefined threshold. In this experiment, we set the threshold to the 50-th percentile of the COMETkiwi scores of MS-Translator, meaning that we use GPT for any translation that has a COMETkiwi score lower than the median MS-Translator COMETKiwi score, which can be easily estimated from previous translation requests.

\autoref{fig:hybrid-gpt} presents the results of our experiments on 12 language pairs. Firstly, we observe that in all language pairs, the ``Hybrid Max-Routing'' approach consistently achieves the highest COMET-22 scores, surpassing both the individual systems. ``Hybrid Max-Routing''  achieves a maximum gain of 1.6 Comet-22 points in the EN-UK language pair which is not among  top performance language for GPT. This indicates that combining the strengths of NMT and GPT systems can lead to a significant improvement in translation quality.

Next, we compare the performance of the individual systems. In general, MS-Translator achieves higher scores than GPT on most language pairs, which is expected given that MS-Translator is an NMT system specifically optimized for translation tasks. However, GPT outperforms MS-Translator on certain language pairs, such as DE-EN, EN-JA, and EN-ZH. This suggests that GPT can be a valuable fallback system in cases where the quality of the primary system is unsatisfactory.

We also compare the performance of the two hybrid approaches. The ``Hybrid Max-Routing'' approach achieves slightly higher scores than the ``Hybrid Threshold'' approach on most language pairs, indicating that routing to GPT only when the quality of MS-Translator falls below a certain threshold may not always be the optimal strategy. However, the ``Hybrid Threshold'' approach still achieves comparable results to the upper bound on all language pairs, while using GPT for only 50\% of the instances. This suggests that it can be a more practical approach in scenarios where computational resources are limited.

\autoref{fig:enxxCOMET22} demonstrates that hybrid approaches achieve larger and more consistent improvements than shots selection across all languages and directions. \autoref{fig:endeCOMET22} zooms in on the high-performing DE-EN and EN-DE systems. The hybrid system outperforms both WMT-Best and MS-Translator systems in both directions, even though the GPT system only outperforms them in DE-EN with 5-shot setup.

In summary, our experiments demonstrate the potential of combining NMT and GPT systems to improve machine translation quality. The results suggest that a hybrid approach that uses GPT as a fallback system can achieve higher performance than either individual systems. Future research can explore more advanced techniques that can leverage the strengths of both systems and optimize the hybrid approach.

\begin{figure*}[h]
  \centering

        \includegraphics[width=0.7\textwidth]{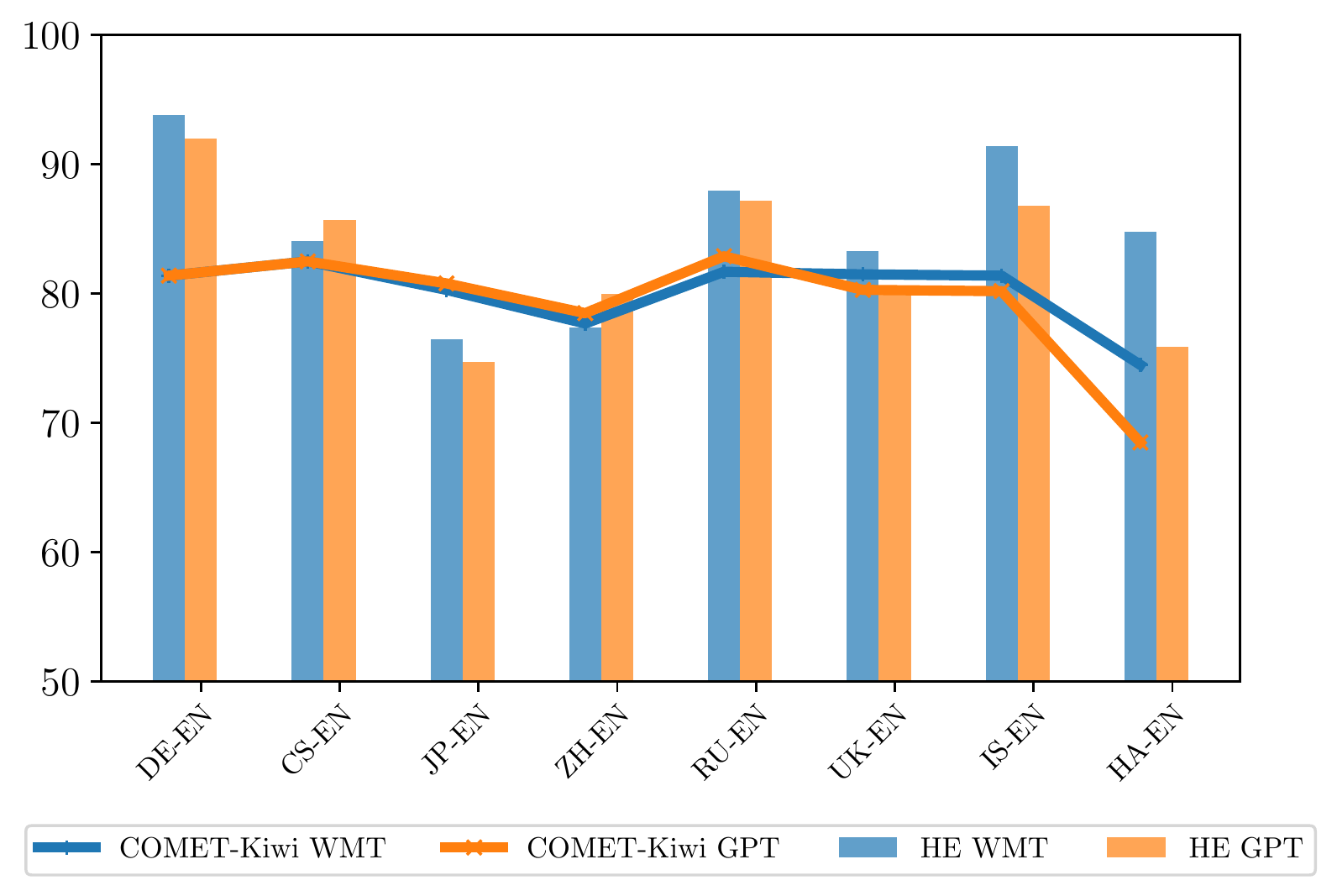}
        \vskip -0.1in
        \caption{\centering Human Evaluation of WMT-Best Systems vs GPT translations for non-English to English}
         \label{fig:HE-COMETkiwi-xe}
   
\end{figure*}

\begin{figure*}[h]
  \centering

        \includegraphics[width=0.7\textwidth]{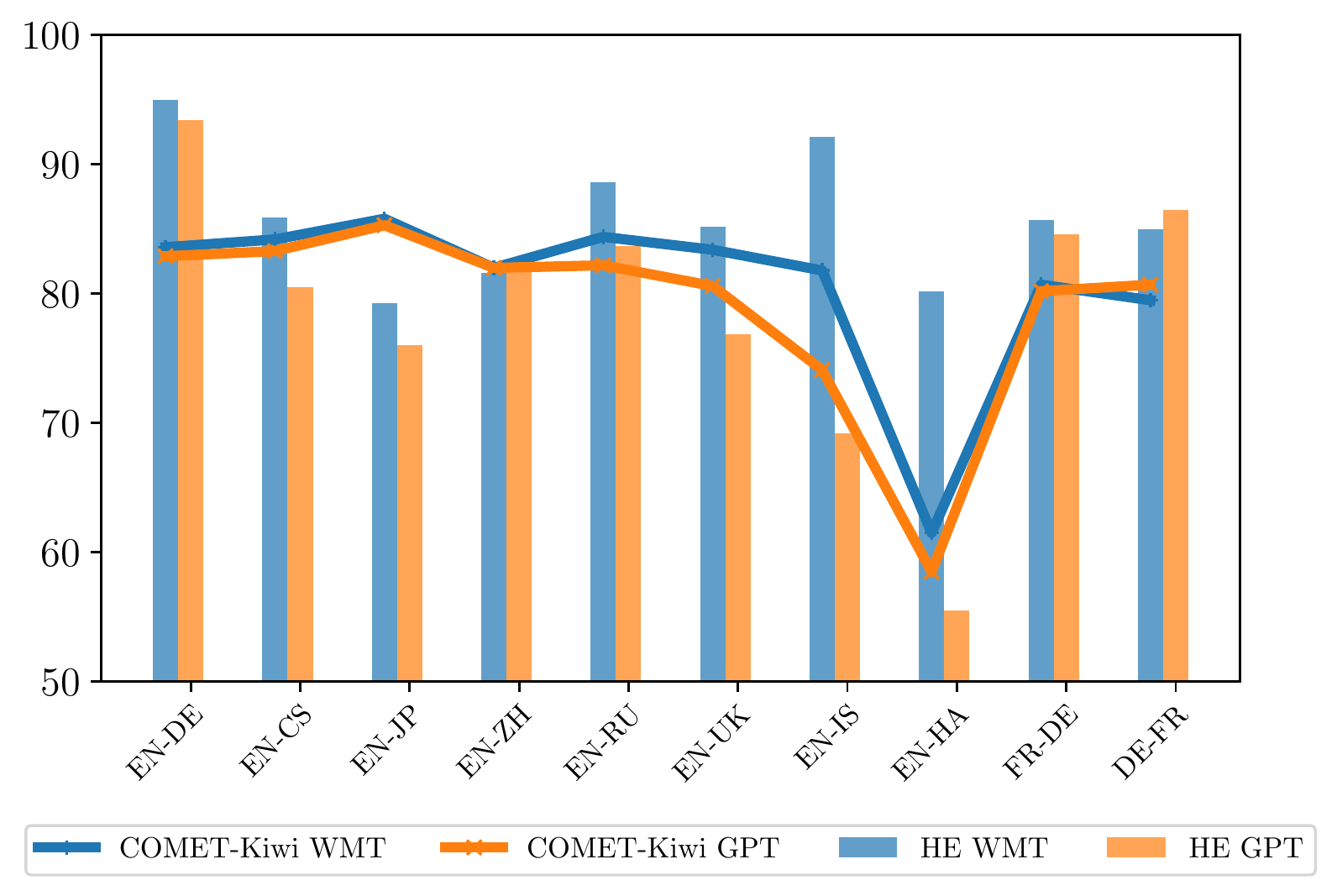}
        \vskip -0.1in
        \caption{\centering Human Evaluation of WMT Best Systems vs GPT translations to non-English}
         \label{fig:HE-COMETkiwi-xx}
   
\end{figure*}

\begin{figure*}[h]
  \centering
        \includegraphics[width=0.7\textwidth]{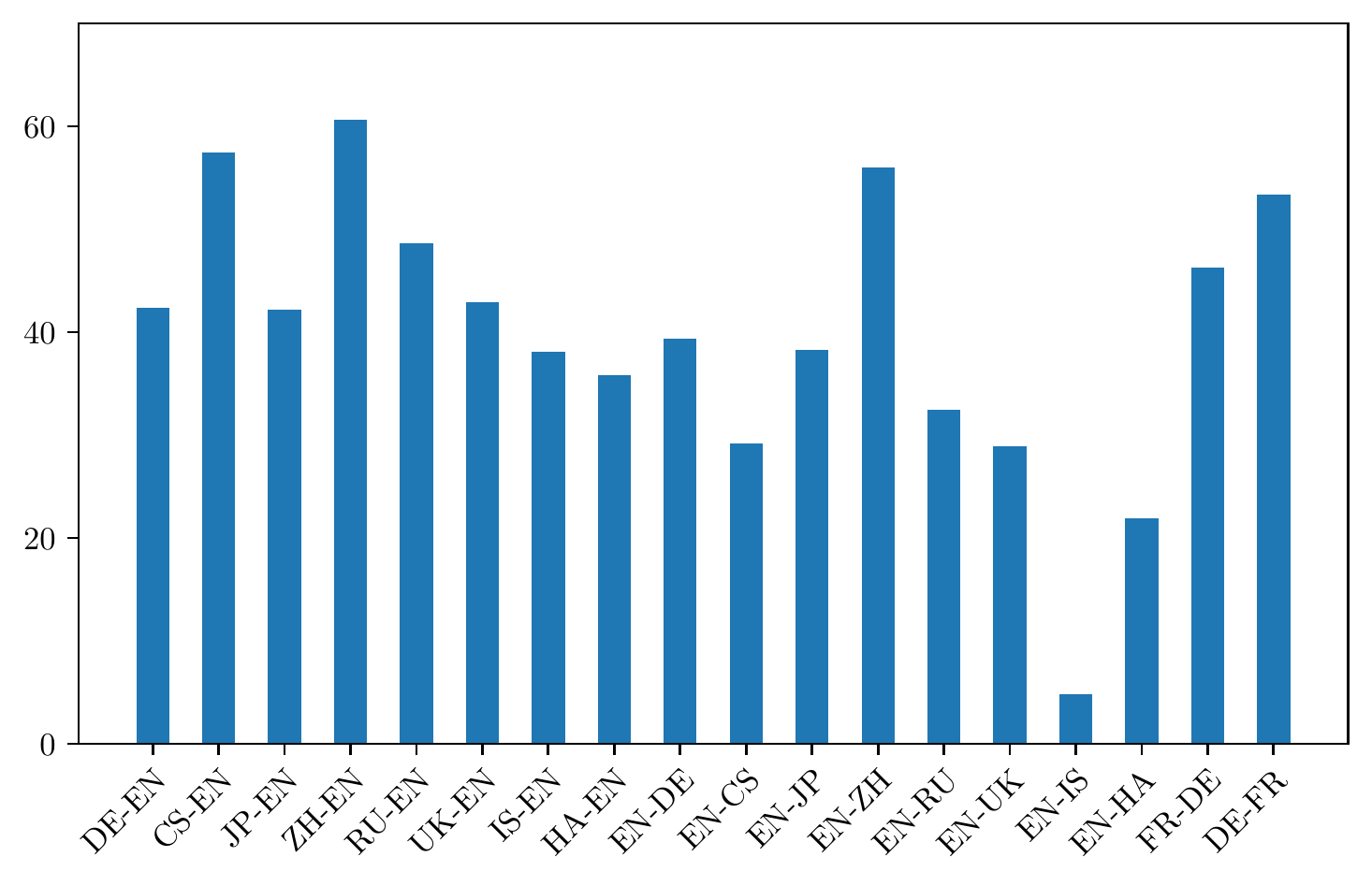}
        \vskip -0.1in
        \caption{\centering Human Evaluation: GPT  Win Rates (\%) based on Item Scores per language pair.}
     \label{fig:gpt-winrate}
\end{figure*}

\section {Human Evaluation and Analysis}
\label{sec:humaneval}

We use source-based sentence-level contrastive Direct Assessment + Scalar Quality Metric (contrastive DA+SQM; \citealt{akhbardeh2021findings}, \citealt{kocmi2022findings}) to perform human evaluation of the WMT-Best systems from~\autoref{tab:datasets} and GPT with 5 shots QR as shown in~\autoref{tab:result.few-full-wmt}. For each language pair, we randomly sample 425 non-identical translation item pairs and have them annotated with the contrastive DA+SQM annotation method by 5 distinct professional translation experts per language pair. \autoref{fig:HE-COMETkiwi-xe} and  \autoref{fig:HE-COMETkiwi-xx} report aggregated human and corresponding COMETkiwi scores. Surprisingly, GPT outperforms Best-WMT systems on CS-EN, ZH-EN, EN-ZH and DE-FR, and achieves comparable results on most of the high-resource languages. On the other hand, the two low-resource languages, Hausa and Icelandic, lag behind significantly. Full details of the scores can be found in  the appendix, \autoref{tab:result.he-qe}.

 We observe that the human evaluation results are highly consistent with the COMETkiwi  results. This highlights the importance of neural reference-less metrics for evaluating MT in general  and this family of models in particular. As we have seen in the previous results, all lexical metrics fail to capture the strong performance of GPT and exhibit lexical and reference bias. While we believe quality estimation is becoming more essential for MT in general, it is reassuring to know that COMETkiwi   performs well on GPT models as well as on NMT models. Moreover, as shown on  \autoref{fig:gpt-winrate}, highly performing GPT languages  pairs demonstrate higher win rate which is reflected on both Human Evaluation results and COMETkiwi scores.

We conducted a manual analysis of human-evaluated GPT translations for English-Japanese and Japanese-English directions to identify their strengths and weaknesses. In~\autoref{tab:he-en-ja-analysis} in the appendix, we present some of the observed characteristics along with examples of GPT and WMT outputs. A notable characteristic is that GPT performs better and more robustly than WMT for source sentences that are erroneous, short, or colloquial. We found that GPT can handle misspellings or unclosed quotation marks and produce translations that do not omit any semantic information. Moreover, GPT can generate reasonable translations for partial or incomplete colloquial source sentences while WMT-Best often adds or omits content. However, GPT tends to produce unnatural translations for sentences with uncommon or complex expressions.

\section {GPT Translation Characteristics}
\label{sec:analysis}

\begin{CJK*}{UTF8}{gbsn}
\begin{table*}[ht]
\centering
\resizebox{2.05\columnwidth}{!}{%
    \begin{tabular}{ l l l}
        \toprule
    \textbf{Sequence Type} & \textbf{Translation Instance} & \textbf{Phenomenon}  \\
        \midrule
Source &  Bis auf die E 95 02 wurden \textbf{alle Lokomotiven zerlegt}. & \\
MS Translator &  With the exception of E 95 02, \textbf{all locomotives were dismantled}. &  \textbf{Non-Monotonicity (NM) } \\
GPT  & \textbf{All locomotives were dismantled} except for the E 95 02. &   \\
\hline
Source &  \textbf{Oder ist sie} ganz aus dem Sortiment genommen? &  \\
MS Translator & \textbf{Or is it} completely removed from the range? &  \textbf{Fluency (F)} \\
GPT &  \textbf{Or has it been} completely removed from the range? &   \\
\hline
Source &  Sehen Sie bitte im Screenshot was der Kollege geschrieben hat &  \\
MS Translator &  Please see in the screenshot what the colleague wrote &  \textbf{Punctuation Insertion (PI)} \\
GPT &  Please see the screenshot for what the colleague wrote\textbf{.} &  \\
\hline
Source &  Die Email zur Stornierung wurde am 26.12.\textbf{\#NUMBER\#} versendet. &  \\
MS Translator &  The cancellation email was sent on 26.12.\textbf{\#NUMBER\#}. &  \textbf{Dropped Content (USW)} \\
GPT  & The cancellation email was sent on December 26th. &  \\
\hline
Source & "We won't accept the \textbf{CAA} and that is for sure.  &  \\
MS Translator & “我们不会接受\textbf{CAA}，这是肯定的。 &  \textbf{Inserted Content (UTW)} \\
GPT &  “我们不会接受《公民法》，这是肯定的。 &   \\
\bottomrule
    \end{tabular}}
    \caption{\textbf{Illustrated Examples of the Phenomena} as described in Section 5. The origin of these differences between translations lie in the computational mechanism leveraged for translations: When controlled for quality, higher translation non-monotonicity suggests a more abstractive computation used for obtaining the translations. Similarly, Fluency, Punctuation Insertion, Dropped and Inserted Content measure different translation characteristics. }
    \label{tab:vikas_examples}
    \vspace{-1.0em}
\end{table*}
\end{CJK*}

\begin{figure}[ht]
\centering
\begin{subfigure}[b]{0.48\textwidth}
\centering
\includegraphics[width=\textwidth]{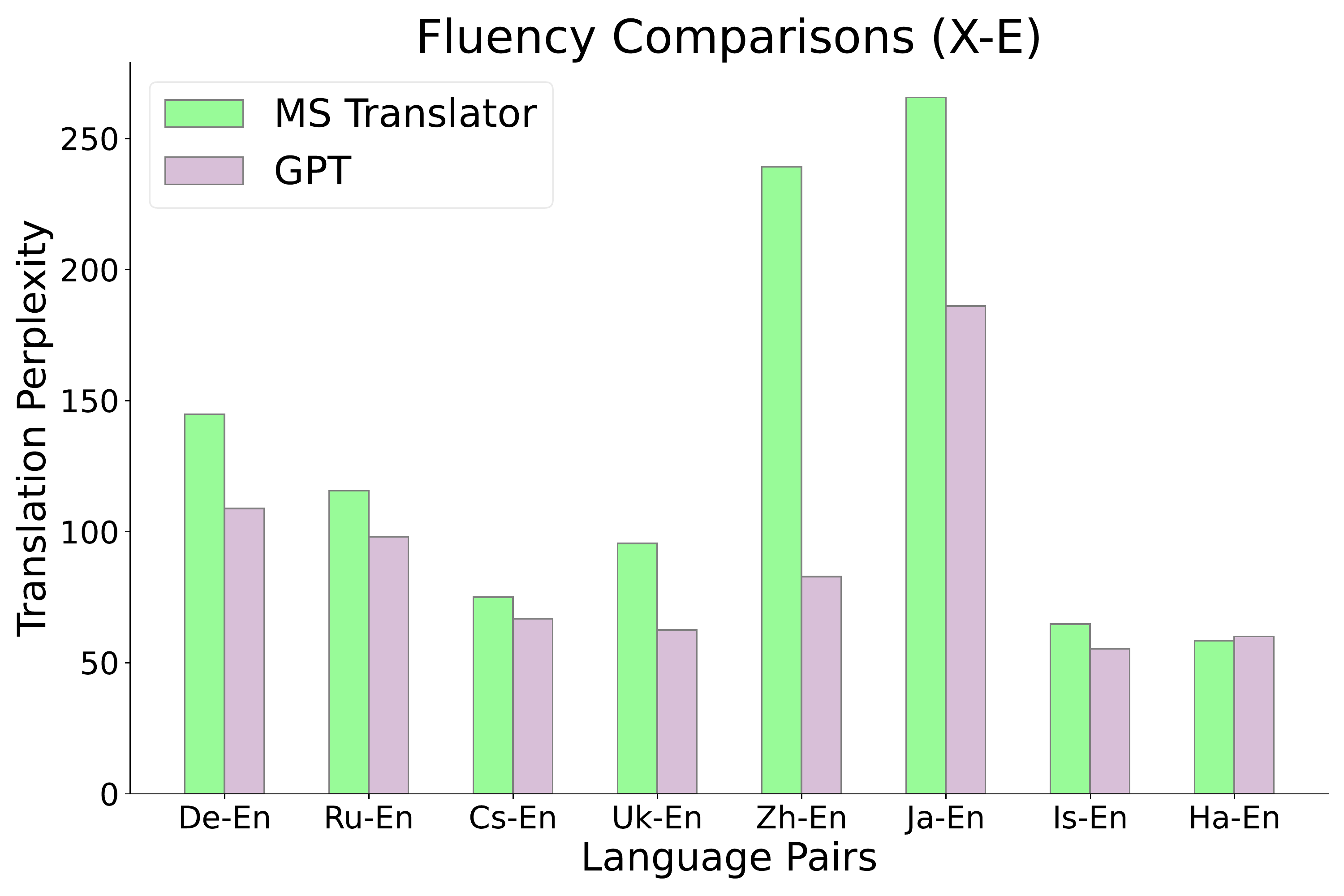} 
\end{subfigure}
\caption{\textbf{Fluency Comparisons} for the \textbf{X-E} language pairs. On 7 out of 8 language pairs, GPT translations obtain lower perplexity, thereby producing more fluent translations. The magnitude of the difference is higher for Zh-En and Ja-En language pairs.}
\label{fig:vikas_plot_1}
\end{figure}

\begin{figure}[ht]
\centering
\begin{subfigure}[b]{0.48\textwidth}
\centering
\includegraphics[width=\textwidth]{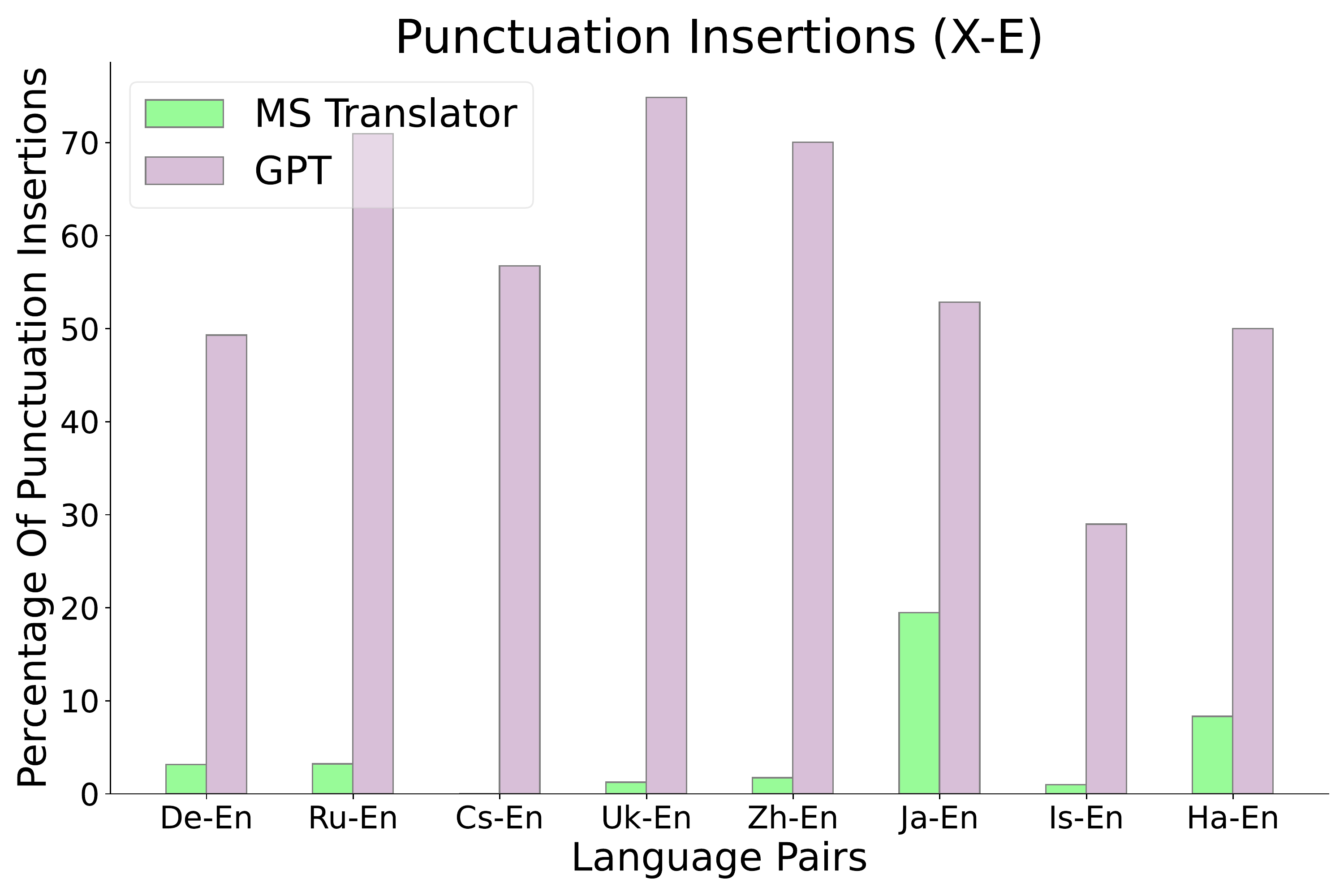} 
\end{subfigure}
\caption{\textbf{Comparisons of Punctuation Insertions} for the \textbf{X-E} language pairs. On 8 out of 8 language pairs, GPT translations show a greater bias towards inserting unsupported end of sentence markers in translation. }
\label{fig:vikas_plot_2}
\end{figure}

\begin{figure}[ht]
\centering
\begin{subfigure}[b]{0.48\textwidth}
\centering
\includegraphics[width=\textwidth]{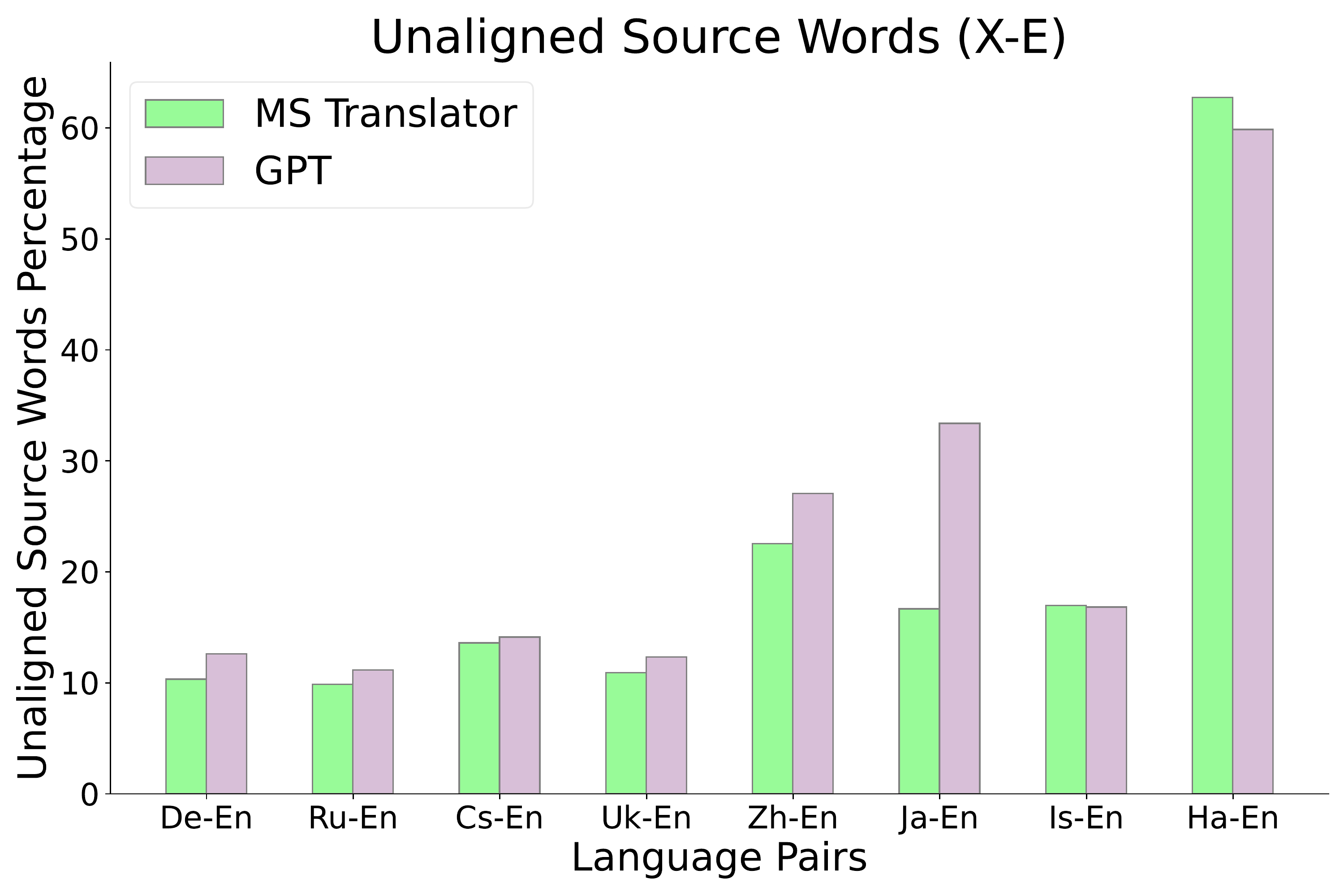} 
\end{subfigure}
\caption{\textbf{Comparisons of Unaligned Source Words} for the \textbf{X-E} language pairs. GPT Translations consistently incur greater number of unaligned source words.}
\label{fig:vikas_plot_3}
\end{figure}

\begin{figure}[ht]
\centering
\begin{subfigure}[b]{0.48\textwidth}
\centering
\includegraphics[width=\textwidth]{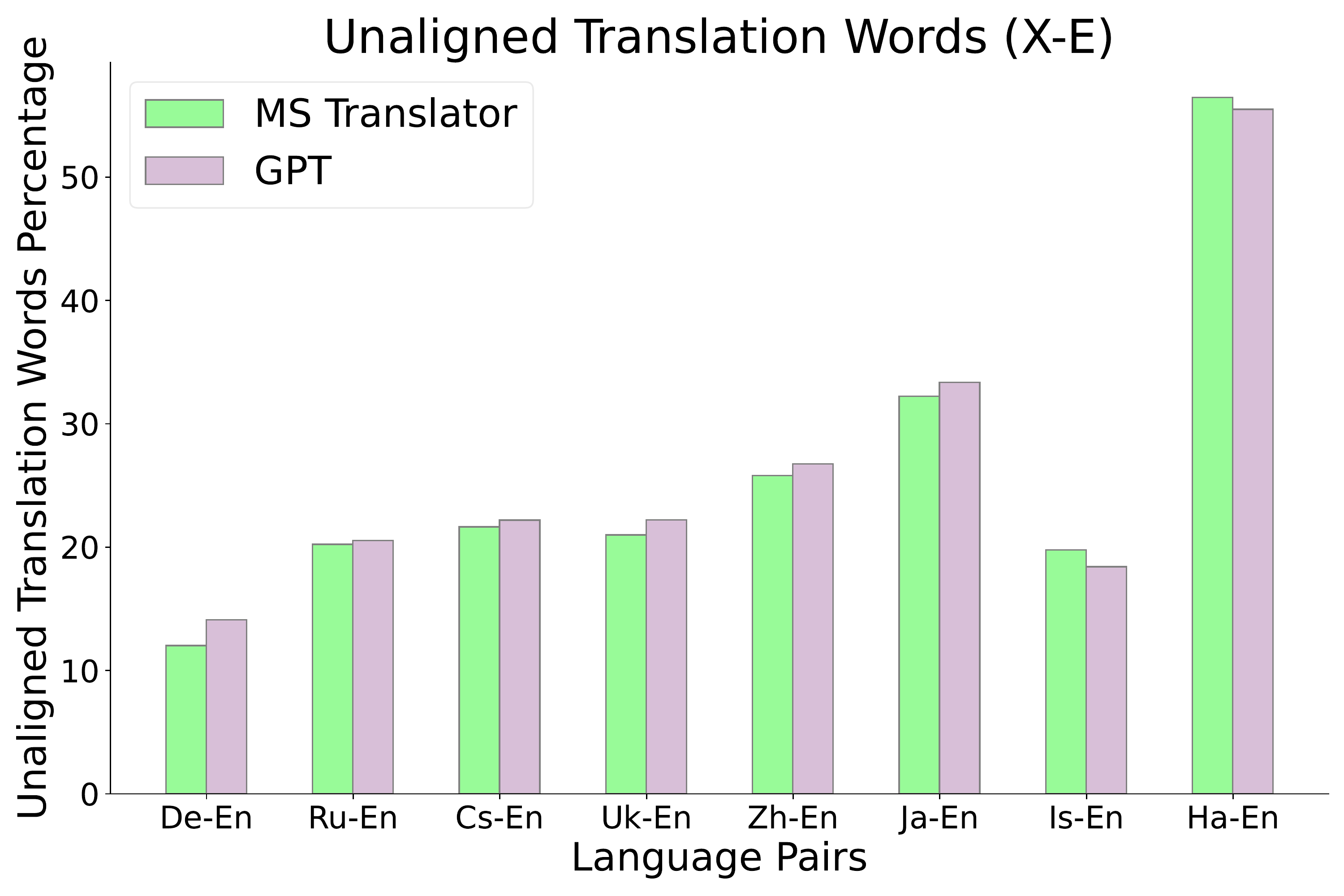} 
\end{subfigure}
\caption{\textbf{Comparisons of Unaligned Translation Words} for the \textbf{X-E} language pairs. GPT Translations consistently incur greater number of unaligned target words.}
\label{fig:vikas_plot_4}
\end{figure}

\begin{figure}[ht]
\centering
\begin{subfigure}[b]{0.48\textwidth}
\centering
\includegraphics[width=\textwidth]{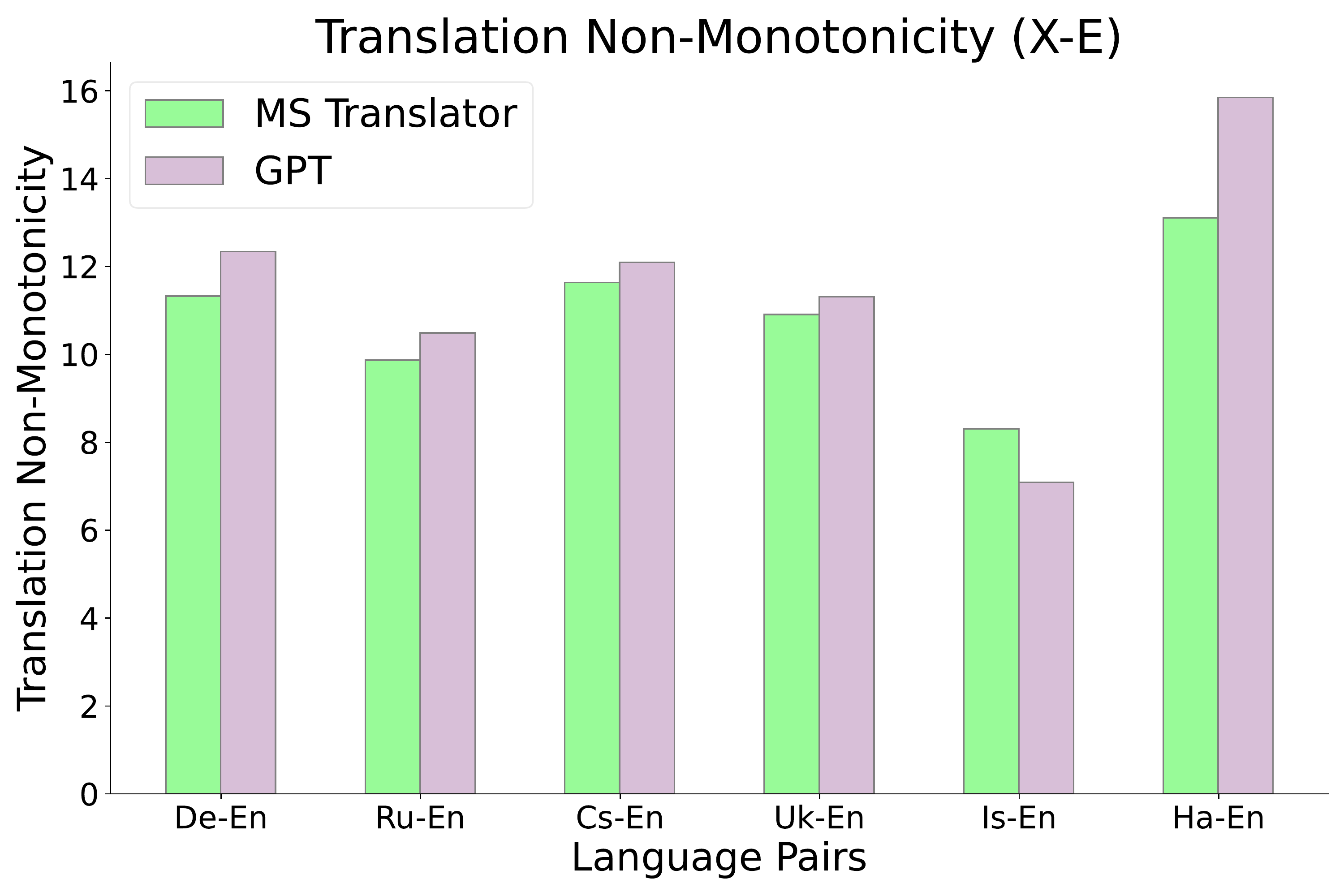} 
\end{subfigure}
\caption{\textbf{Comparisons of Translation Non-Monotonicity} for the \textbf{X-E} language pairs. GPT Translations consistently score higher on the non-monotonicity of translations.}
\label{fig:vikas_plot_5}
\end{figure}

\begin{figure}[ht]
\centering
\begin{subfigure}[b]{0.48\textwidth}
\centering
\includegraphics[width=\textwidth]{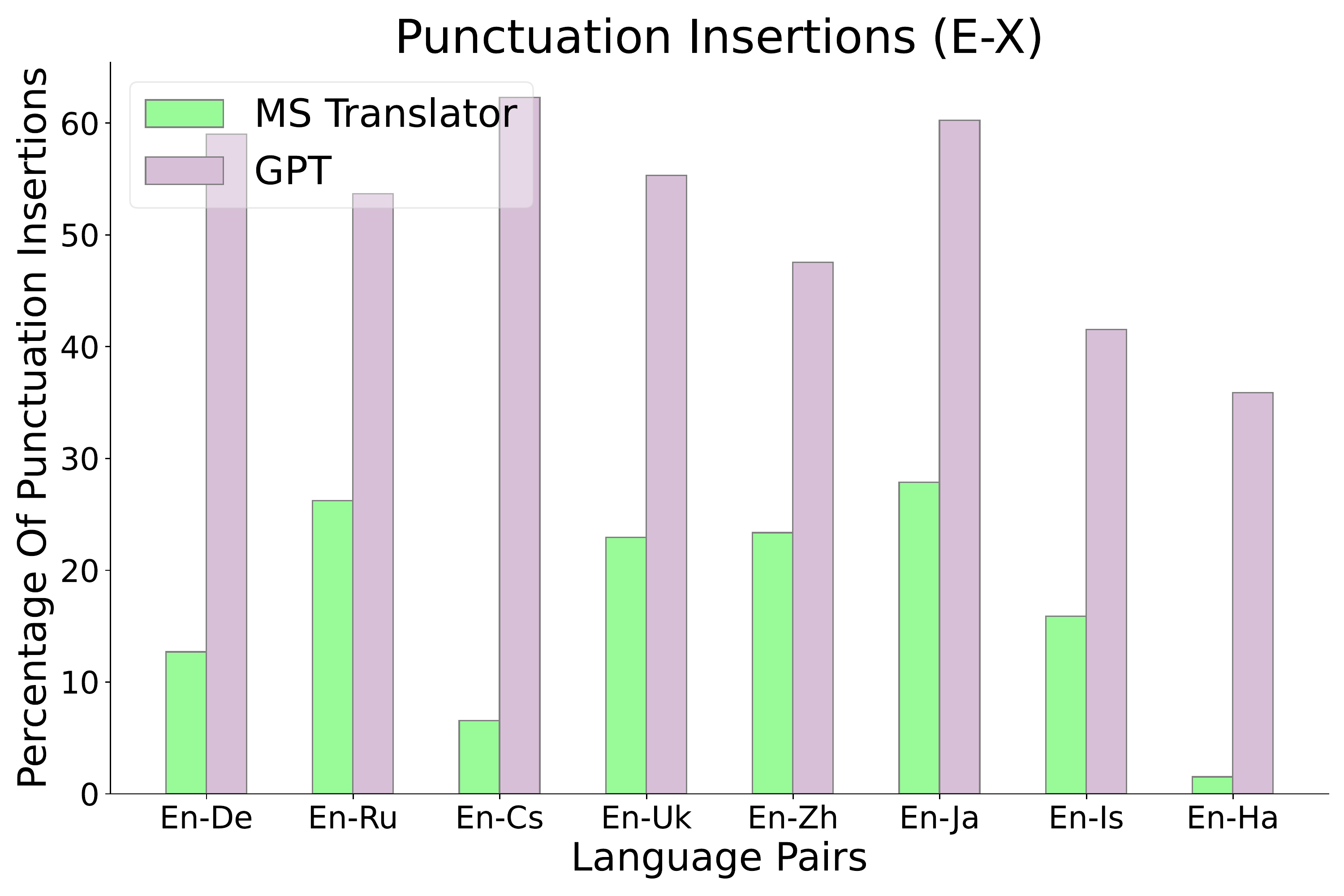} 
\end{subfigure}
\caption{\textbf{Comparisons of Punctuation Insertions} for the \textbf{E-X} language pairs. On 8 out of 8 language pairs, GPT translations obtain higher scores.}
\label{fig:vikas_plot_6}
\end{figure}

\begin{figure}[ht]
\centering
\begin{subfigure}[b]{0.48\textwidth}
\centering
\includegraphics[width=\textwidth]{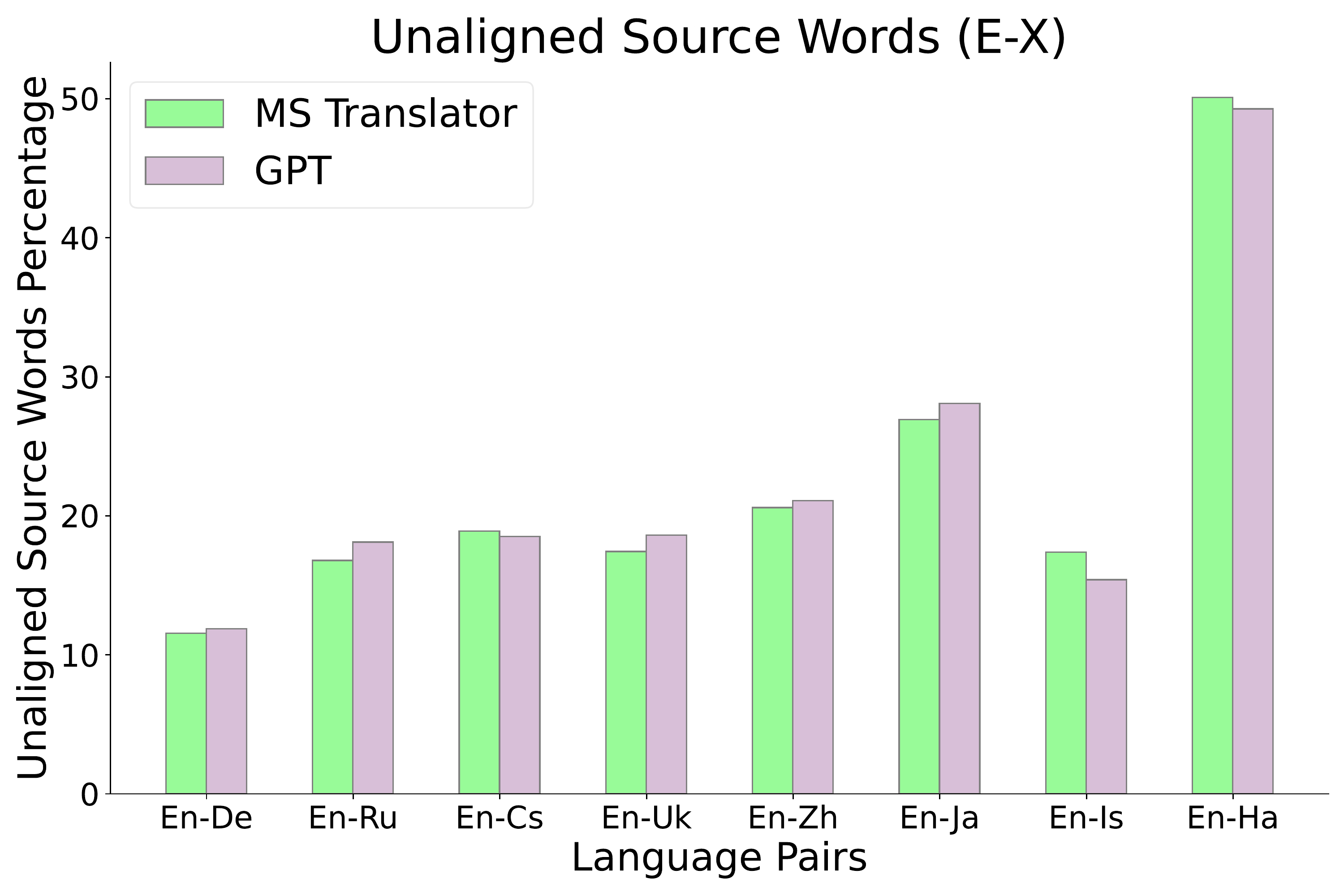} 
\end{subfigure}
\caption{\textbf{Comparisons of Unaligned Source Words} for the \textbf{E-X} language pairs. GPT Translations, on average, incur greater number of unaligned source words.}
\label{fig:vikas_plot_7}
\end{figure}

\begin{figure}[ht]
\centering
\begin{subfigure}[b]{0.48\textwidth}
\centering
\includegraphics[width=\textwidth]{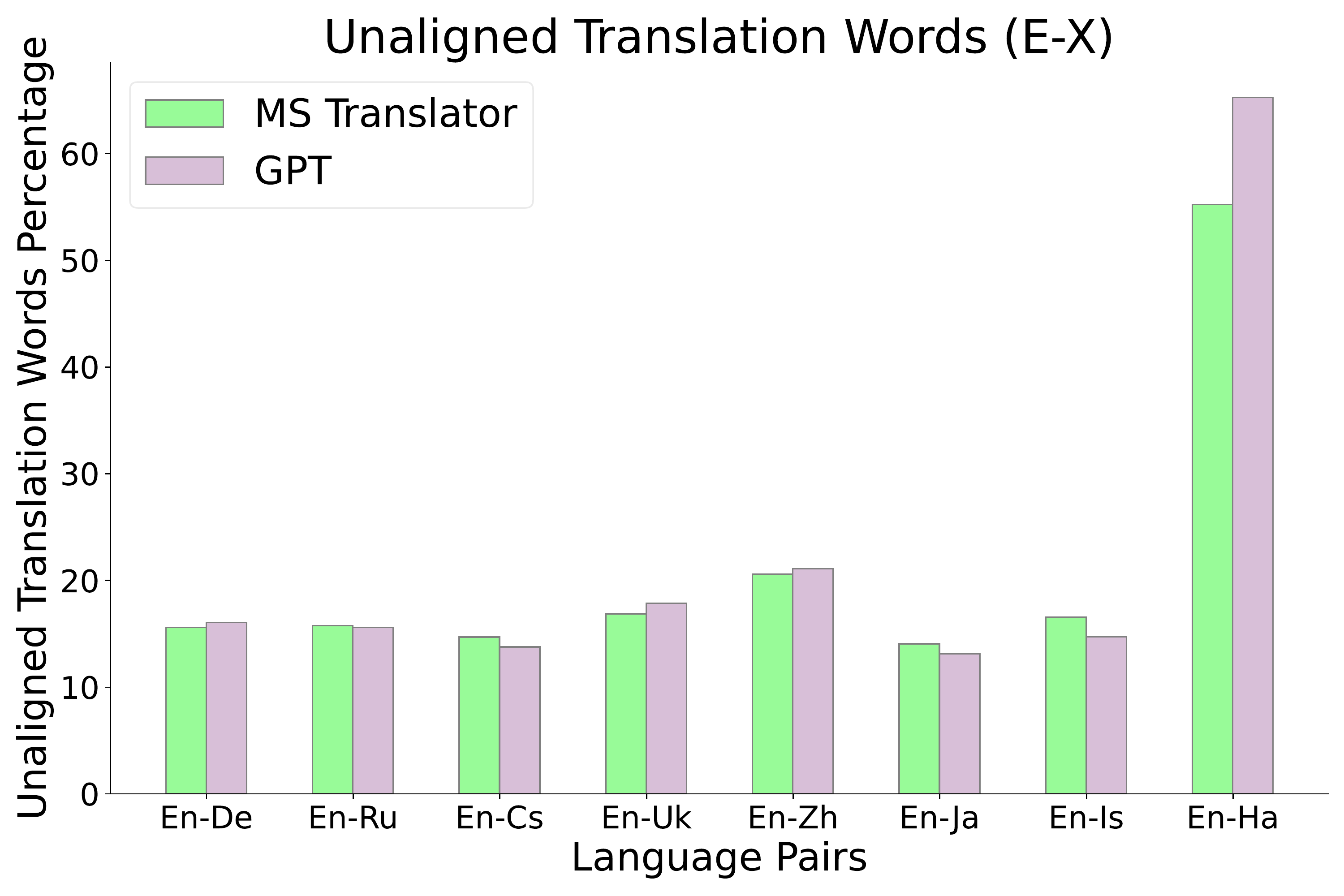} 
\end{subfigure}
\caption{\textbf{Comparisons of Unaligned Translation Words} for the \textbf{E-X} language pairs. GPT Translations consistently incur greater number of unaligned target words.}
\label{fig:vikas_plot_8}
\end{figure}

\begin{figure}[ht]
\centering
\begin{subfigure}[b]{0.48\textwidth}
\centering
\includegraphics[width=\textwidth]{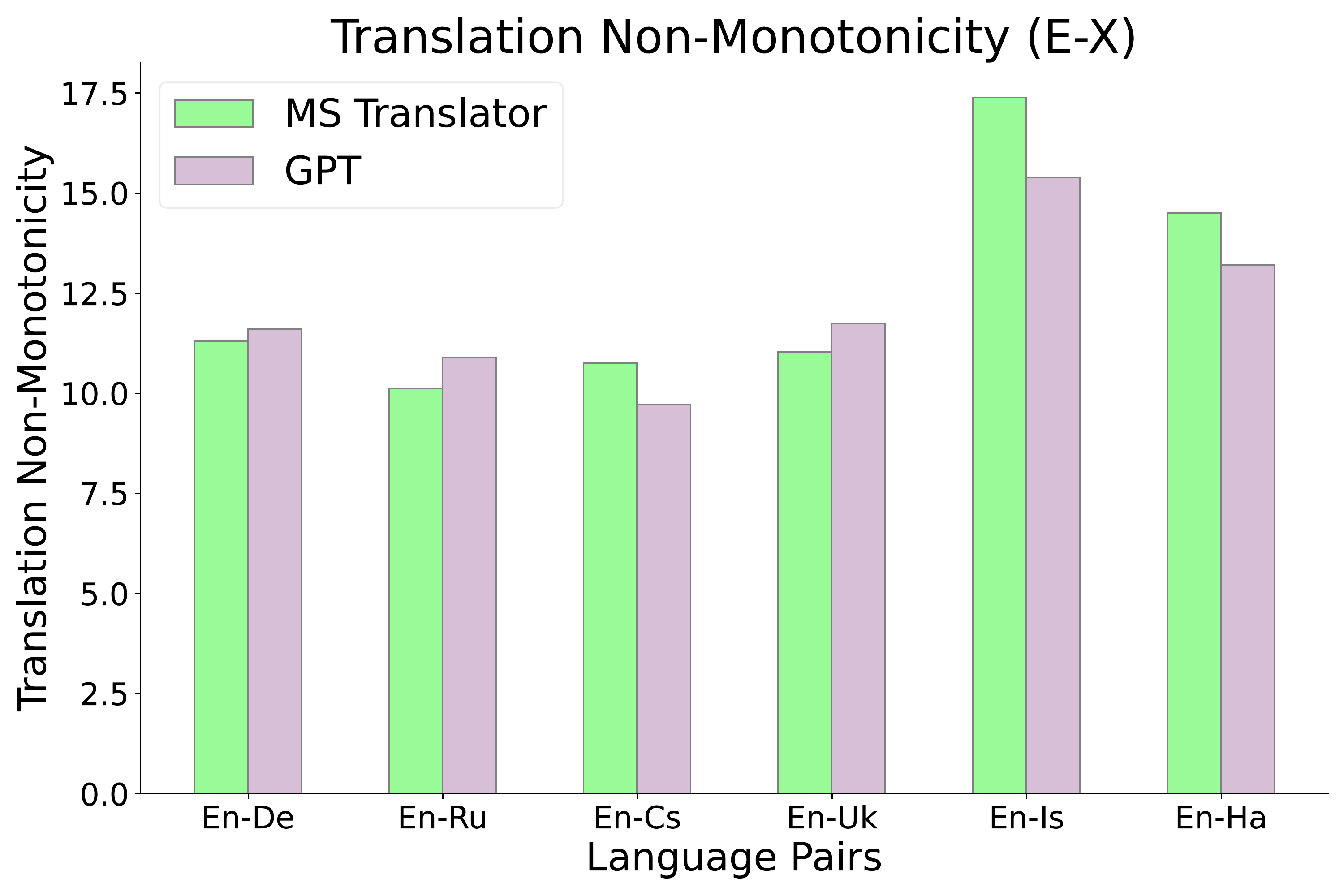} 
\end{subfigure}
\caption{\textbf{Comparisons of Translation Non-Monotonicity} for the \textbf{E-X} language pairs. GPT Translations score higher on translation non-monotonicity for 3 out of 4 high-resource language pairs.}
\label{fig:vikas_plot_9}
\end{figure}

In this section, we try to comprehensively analyze the characteristics of the translations obtained from GPT. Our goal here is to better differentiate GPT translations from its NMT counterpart.

\subsection{Situating GPT Translations}
\label{sec:analysis1}
We posit that there are two key biases due to which the computation of translation done by LLMs might be different from the same computation done by NMT models, namely the \textit{Parallel Data Bias} and the \textit{Language Modeling Bias}.\\

\paragraph{Parallel Data Bias:} Compared to NMT models trained on parallel data, which is typically web-mined (and noisy), LLMs such as GPT are trained on monolingual data only with no explicit supervisory signal for the translation task. This creates interesting implications on the nature of the emergent computational abilities leveraged for our task of interest, translation. First, not using parallel data might imply that LLMs are protected against the noise associated with parallel data, which leads to problems such as the memorization of noisy/atypical \cite{naacl_hallucinations} or low-quality samples \cite{emnlp_low_quality_data} and biases towards particular language characteristics predominant in the parallel data \cite{garcia2023unreasonable}. These \textit{parallel data biases} can also manifest in the form of long-tailed errors such as the translations of physical units or currencies \cite{emnlp_long_tail_errors}, owing to a preponderance of such incorrect token pairings in the parallel data. On the other hand, the lack of explicit supervisory signals for the task could also mean that LLM based translations might not track the desired characteristics of translations such as faithfulness to the source as well as the NMT models trained with explicit teacher-forced supervision \cite{anonymous2023literalness}.

\paragraph{Language Modeling Bias:} Despite the impressive performance of in-context learning, constraining LLM behavior to explicitly follow the specifications of a desired task is a non-trivial problem. Analyses of in-context learning have revealed how the \textit{implicit} zero-shot performance of LLMs might be higher than their \textit{observed} zero-shot performance, with the demonstrations within in-context learning themselves providing only limited learning signals \cite{min2022rethinking, kojima-zero-shot, anonymous2023dissecting}. A direct implication of these results for translation is that the demonstrations used for in-context learning might \textit{fail} to override the underlying computational bias of language modeling which is likely to favor greater fluency at the cost of adequacy. Such \textit{language modeling bias} might also introduce undesirable artifacts, e.g., punctuation insertions, acronym expansions, world knowledge insertion, etc. in the translations which could cause it to veer off from a faithful cross-lingual representation of the input.


In the next subsection, we propose properties along which finer-grained characteristics of GPT translations could be enumerated. These measures are designed to provide indirect measurements of the language modeling bias as well as the parallel data bias, which could allow a better differentiation of GPT translations against translations from NMT systems. We first discuss the measurements designed to elicit artifacts associated with the language modeling bias.


\subsection{Language Modeling Bias Artifacts}
\label{sec:bias}
We propose and use five measurements over the test sets to quantitatively explore language modeling bias, in order to enumerate the differences in translations obtained from traditional NMT systems and GPT. Below, we describe the properties as well as the algorithms used for quantifying them (corresponding illustrative examples of the phenomena are presented in~\autoref{tab:vikas_examples}):

\begin{enumerate}
    \item \textbf{Translation Non-Monotonicity} (\textbf{NM}): We aim to measure how closely the translation tracks the source sentence. A more paraphrastic or a less literal translation is likely to deviate from a close tracking of the source word order (across language pairs). We use the non-monotonicity metric proposed in \citet{additive_interventions}, which computes the deviation from the diagonal in the word to word alignment as the non-monotonicity measure. This measurement could also be interpreted as a normalized measure of alignment crossings, which has been shown to correlate with translation non-literalness \cite{literalness_measuring}. This measurement has also been used in \citet{anonymous2023literalness} for investigating translation literalness. 
    \item \textbf{Translation Fluency} (\textbf{TF}): We measure translation fluency using a strong, independently trained language model (`gpt2-large', \citet{gpt2}). We restrict this measurement to X-E direction, since GPT-2 has only been trained on English text \cite{gpt2}.
    \item \textbf{Punctuation Insertion} (\textbf{PI}): Language modeling bias can prefer one mode of sentence completion in contrast to others. This can reveal itself in the presence of not well-formed inputs such as sentences that do not end with typical end of sentence markers (comma, period and exclamation). We measure the fraction of input sentences for which the translation contains an end of sentence marker but the source does not. The insertion of an end of sentence marker in such instances is inadequate for translation, a task which strives for bitext equivalency.
    \item \textbf{Unaligned Source Words} (\textbf{USW}): We measure the number of source words left unaligned in a word to word alignment obtained over the source and output translations. When controlled for quality, a more paraphrastic translation is likely to contain more words that do not align with the words in the source sentence. This measurement was used in \citet{anonymous2023literalness} as a measure of translation literalness and we use it similarly to obtain a measurement of content that is dropped in a translation -- an untranslated word or phrase in a source sentence is likely to find no alignments in the output. For obtaining word to word alignments, we use a multilingual-bert based aligner \cite{devlin-etal-2019-bert, awesome_aligner}.
    \item \textbf{Unaligned Translation Words} (\textbf{UTW}): We measure the number of unaligned words in the translation using the same word to word alignments as in the previous measurement. This indicates the presence of words that have no support in the source and is included to measure words that are potentially inserted in the translation without any basis in the input.
\end{enumerate}

We collect the measurements on these properties over the test sets for all the language pairs under investigation. We compare MS Translator with GPT throughout. We report the results in the next section and present our analysis grouped by the translation direction.

\subsection{X-E Translation characteristics}
\label{sec:xe}

Figures \ref{fig:vikas_plot_1}, \ref{fig:vikas_plot_2}, \ref{fig:vikas_plot_3}, \ref{fig:vikas_plot_4} and \ref{fig:vikas_plot_5} represent the comparisons of GPT translations against MS Translator for X-E language pairs. \autoref{fig:vikas_plot_1} shows that GPT translations obtain lower perplexities, thereby demonstrating greater fluency. \autoref{fig:vikas_plot_2} shows that GPT translations suffer from the problem of punctuation insertion with a much higher frequency than MS translator. We attribute this to the language modeling bias, which would prefer to generate a well-formed sentence, even if such well-formed-ness is unsupported in the input. \autoref{fig:vikas_plot_3} shows that GPT translations incur slightly higher number of unaligned source words on 7 out of 8 eight language pairs. Greater unaligned source words would imply either the presence of greater paraphrasticity in the translations or greater inadequacy (dropped or inserted content). \autoref{fig:vikas_plot_4} shows that GPT translations incur almost similar number of unaligned target words, suggesting that the GPT translations are similarly adequate in terms of potential insertions. Another measurement, presented in \autoref{fig:vikas_plot_5} shows that GPT translations are more non-monotonic than its NMT counterpart.

\subsection{E-X Translation characteristics}
\label{sec:ex}

Figures \ref{fig:vikas_plot_6}, \ref{fig:vikas_plot_7}, \ref{fig:vikas_plot_8} and \ref{fig:vikas_plot_9} represent the comparisons of GPT translations against MS Translator for E-X language pairs. \autoref{fig:vikas_plot_6} shows that similar to X-E translations, GPT E-X translations also suffer from a higher frequency of punctuation insertions. The magnitude of the difference however, is smaller than the X-E translations, suggesting a weaker language modeling bias for these languages. \autoref{fig:vikas_plot_7} shows that in general, GPT translations incur greater number of unaligned source words than its NMT counterpart. \autoref{fig:vikas_plot_8} shows that the number of unaligned translation words for GPT translations do not differ greatly from MS Translator. Similarly, \autoref{fig:vikas_plot_9}, which compares translation non-monotonicity shows no aggregate trends. As such, we find that the translation characteristics for E-X language language pairs depends heavily on the individual language pair under consideration.

%






\begin{table}
  \centering
  \small
\setlength\tabcolsep{2.0pt}
  \begin{tabular}{llcccc}
    \toprule
    \textbf{Lang-Pair} & \textbf{System} & \textbf{PI $\downarrow$} & \textbf{NM $\downarrow$} & \textbf{USW $\downarrow$} & \textbf{UTW $\downarrow$} \\
    \midrule
   De-Fr & MS-Translator & 3.61 & 17.68 & 10.91 & 25.39 \\
    & GPT & 42.98 & 17.21 & 11.42 & 25.11 \\ \midrule
   Fr-De & MS-Translator & 2.63 & 14.63 & 21.87 & 14.63 \\
    & GPT & 60.30 & 14.52 & 21.77 & 14.52 \\
    \midrule
  \end{tabular}
  \vspace{-0.75em}
  \caption{Translation Characteristics Comparisons for\textbf{ De-Fr and Fr-De}: GPT shows a much higher tendency to add an end of sentence marker into the translation when it is absent in the source.}
  \label{tab:vikas_direct}
\end{table}

\subsection{X-Y Translation characteristics}
\label{sec:xy}

\autoref{tab:vikas_direct} reports the results of the five measurements for De-Fr and Fr-De translation directions. The results for direct translation pairs are quite different from the X-E and E-X cases, since typically non-English centric translations are done through pivoting. As such, the trends for measurements over Fluency (F), Unaligned Source Words (USW), Unaligned Translation Words (UTW) and Translation Non-Monotonicity (NM) do not show any conclusive evidence of greater paraphrasticity of GPT translations. However, GPT translations still produce greater number of punctuation insertions than the MS Translator system.






\subsection{Parallel Data Bias Artifacts}
\label{sec:parallel}

To illustrate the parallel data bias, we analyze the translations on low-quality inputs. The intuition behind our experiment is that low-quality inputs are more likely to correspond to the noisy parts of the parallel data that underlie NMT systems trained on large-scale datasets mined from the web. As such, GPT should outperform NMT systems on such low-quality inputs.

\paragraph{Experiment:} We split the test sets into 3 buckets based on perplexity of the source sentence. We notice that the highest perplexity inputs typically correspond to ill-formatted texts, with many inputs pertaining to the e-commerce domain. Such inputs are more likely to \textit{resonate} with the noisy parts of the parallel corpora on which NMT models are typically trained. For example, such high-perplexity inputs might correspond to ill-formed texts scraped from e-commerce websites usually present in parallel corpora. Since we use GPT-2 for obtaining perplexities, we conduct this experiment for E-X language pairs only.

\paragraph{Results:} \autoref{tab:vikas_ppl} presents the results of the experiment across different language pairs. The measurement reported is the average difference in the quality between GPT and MS Translator as measured using COMET-KIWI. We observe that on English to Chinese, English to Japanese and English to Russian language pairs, on which parallel data mining is typically harder owing to a change in script, GPT translations obtain higher performance than MS Translator on the highest perplexity bucket. For low-resource language pairs, GPT gains proportionally in even the lowest perplexity buckets. 

Overall, we find that that in four out of five high resource language pairs in \autoref{tab:vikas_ppl}, GPT obtains higher improvements in the bucket with the highest input perplexity, when compared to the other lower-perplexity buckets. In the cases of English (Latin script) to Chinese, English to Japanese and English to Russian (Cyrillic script), the differences follow a monotonic order with respect to input perplexity. The results suggest that for these language pairs, GPT does obtain better performance on the lower quality inputs. We attribute this behavior to the parallel data bias. Such parallel data noise biases are likely to be correlated with input domains as well, but we leave such an exploration to future work. \\

\begin{table}
  \centering
  \small
\setlength\tabcolsep{1.50pt}
  \begin{tabular}{lccccccc}
    \toprule
    \textbf{Split} & \textbf{En-De} &\textbf{En-Ru} & \textbf{En-Cs} & \textbf{En-Zh} & \textbf{En-Ja} & \textbf{En-Is} & \textbf{En-Ha}\\
    \midrule
    Lowest & \textbf{-0.02} & -0.68 & \textbf{-0.43}  & 0.22 & -0.04 & \textbf{-6.34} & \textbf{1.63} \\
    Medium & -0.48 & -0.41 & -0.97  & 0.47 & -0.18 & -5.71 & 1.21 \\
    Highest & -0.18 & \textbf{-0.14} & -1.10 & \textbf{1.63} & \textbf{0.38} & -6.15 & 1.11 \\ \midrule
  \end{tabular}
  \vspace{-0.75em}
  \caption{\textbf{Exploring Parallel Data Bias}: On English to Chinese, Japanese and Russian language pairs, GPT translations obtain higher performance than MS Translator in the highest perplexity bucket. For low-resource language pairs, GPT gains proportionally in even the lowest perplexity buckets.}
  \label{tab:vikas_ppl}
\end{table}




\subsection{Summary}

We demonstrated that the computational mechanisms operating behind LLMs and NMT models produce translation artifacts that can be quantitatively differentiated. In this subsection, we summarize our comprehensive characterization of the translations produced by GPT.  

\paragraph{Improvements Produced by GPT:} For X-E translations, the translations produced by GPT are more fluent, obtaining consistently lower perplexity (as shown in \autoref{fig:vikas_plot_1}). At the same time, GPT translations for X-E language pairs generally incur higher number of unaligned source words (\autoref{fig:vikas_plot_3}) and in general, similar number of unaligned target words (\autoref{fig:vikas_plot_4}). GPT translations are also more non-monotonic, producing translations that involve longer range reorderings (\autoref{fig:vikas_plot_5}). The combination of these results yields an interesting conclusion: that X-E translations by GPT are more fluent and more paraphrastic than the NMT system under investigation (MS Translator), while being faithful to the source. The greater paraphrasticity is not accompanied by content that is unsupported by the source, i.e., the problem of inserted factual content is not a prominent issue in these language pairs.

For E-X translations, GPT incurs a greater number of unaligned source words (USW, \autoref{fig:vikas_plot_8}), along with greater translation non-monotonicity in general (NM, \autoref{fig:vikas_plot_9}), suggesting greater paraphrasticity. However, at the same time, GPT translations incur a slightly higher number of unaligned translation words as well. This suggests that greater paraphrasticity is not the only cause behind the higher USW and NM measurements, and a \textit{less adequate} translation than the NMT system under investigation is a plausible cause behind these observations as well. This is corroborated by the lower quality measurements for E-X GPT translations obtained previously. In general, we find that for deriving conclusions about GPT translation quality for E-X, it is more important to focus on the single language pair under consideration, i.e. the target non-English language is of critical importance and the effects of language modeling bias cannot be generalized as in the case of X-E translations.


\paragraph{Areas of Improvements:} One artifact of the language modeling bias is that GPT inserts end of sentence markers not present in the source with a far greater frequency than the NMT system under investigation. This holds true across both X-E, E-X and X-Y translation directions. Such a proclivity towards greater fluency might not be appropriate for domains wherein a very literal (and faithful) translation is desired. Similarly, greater paraphrasticity might not be appropriate for certain domains. Also, a related area of improvement for future evaluations would be to conduct separate evaluations of fluency and adequacy, in addition to joint adequacy and fluency quality estimation done presently. Instituting a norm of using multi-dimensional automatic quality measurements \cite{emnlp_long_tail_errors} can provide very targeted signals on differentiating aspects of translation quality, useful especially when there are competing state-of-the-art approaches.

\paragraph{Areas of Application:} Our results also suggest that the greater paraphrastic nature of GPT translations could have applications in improving NMT models on the translation of figurative text. Similarly, the greater gains obtained by GPT translations in the highest perplexity buckets of multiple E-X the test sets suggest that GPT translations might be favored over NMT models when the input domain is likely to contain noisy, ill-formed sentences. These two application areas, based on the demonstrated characteristics of GPT translations, offer two avenues that could benefit from a combination of NMT models with GPT. For example, based on the results in~\autoref{tab:vikas_ppl}, a hybrid English-Japanese system that could improve upon both MS Translator and GPT translations would be the one wherein the highest perplexity inputs are routed to be translated by GPT whereas the lower perplexity inputs are translated through NMT models (e.g., MS Translator). Such a composition might be able to leverage \textit{complementary} strengths of NMT and LLM systems for translation.






\section{Multilingual Capabilities beyond Translation}
\label{sec:beyond}

In this section, we investigate the multilingual capabilities of GPT models beyond translation. Specifically, we aim to assess how well the models perform on emerging reasoning tasks\footnote{ as recently studied in the chain of thought paradigm} for various languages compared to English. We are interested in understanding the degree of multilingual support that GPT models can offer given their translation performance. That is, can we use the translation performance as a proxy for the multilingual performance on other tasks?

We use MGSM Benchmark\cite{mgsm} which is a  Multilingual Grade School Math (MGSM) arithmetic reasoning benchmark. The multilingual problems are human translated from the English dataset GSM8K which is English-language human-annotated grade-school math problem dataset. The dataset supports a set of ten languages other than English (EN): Bengali (BN), Chinese (ZH),
French (FR), German (DE), Japanese (JA), Russian (RU),
Spanish (ES), Swahili (SW), Telugu (TE), and Thai (TH).

\begin{table*}[!htbp]
\centering
\vspace{-0.9em}
\scalebox{1.0}{
\begin{tabular}{c|  c |  c c c| c c c | c c c c }
\hline
\textbf{Setup} &   \textbf{EN} &
\textbf{DE}  & \textbf{FR} & \textbf{ES}  & \textbf{RU}  &  \textbf{ZH}  & \textbf{JA}  & \textbf{TH}  & \textbf{TE}  & \textbf{BN}  & \textbf{SW} \\ \hline

& \multicolumn{11}{c}{GPT text-davinci-002}  \\ 
Native-CoT &  	53.6 & 36.0 & 37.6 & 40.4 & 28.4 & 40.0  & 26.0  & 10.8 & 0.4 & 6.4 &11.2 \\
Translate-EN &  53.6 & 46.4 &46.4 & 51.6 & 48.8 & 47.2 & 44.8 & 41.2 & 42.8 & 41.2 & 37.6 \\\hdashline

& \multicolumn{11}{c}{GPT text-davinci-003}  \\ 
Native-CoT &  	56.8 & 54.8 & 53.6 &	59.2  &	40.8  
& 44 & 38.8 & 25.9 &7.0 & 14.1 & 15.2 \\
Translate-EN &  56.8 & 51.6 & 50.5  & 49.2 &	54  & 48.4 & 46.7  &  31.2 &  44.8 & 47.6   & 45.76 \\
Translate-EN+ &  56.8 & 49.6  & 53.2  & 58.8  &	57.6  & 49.2 & 46.8  & 30.8  &  42.5 &  48.1  & 44.9 \\\hdashline

& \multicolumn{11}{c}{PaLM-540B}  \\ 
Native-CoT &  62.4 & 49.2 & 46.4 & 56.8 & 48.4 & 46.8 & 40.0 & 52.8 & 45.6 &46.0 &35.2 \\
Translate-EN &62.4 & 57.2 & 55.2 & 60.0 & 59.6 & 55.6 & 50.0 & 50.8 &49.6 & 53.2 &51.2 \\\hline

\end{tabular}}
\caption{GPT performance on MGSM dataset. PaLM-540 results from\cite{mgsm}.}
\label{tab:result.mgsm}
\vspace{-1.5em}
\end{table*}

\autoref{tab:result.mgsm} presents the results on the MSGM benchmark. We first use Native-CoT, which uses prompts and CoT in the native language of each dataset. We observe that \emph{text-davinci-003} surpasses \emph{text-davinci-002} for all languages, highlighting the effectiveness of \emph{text-davinci-003} on multilingual tasks. The performance is especially high on EN, DE, FR and ES, while RU, JA and ZH exhibit lower scores than the Latin languages. The low-resource languages, however, achieve limited performance, indicating the need for better approaches to attain truly multilingual support.

We then use Translate-EN, which translates all prompts and CoT into English. We find that this setup enhances the performance on the non-Latin group (RU, JA and ZH) as well as the low-resource group (TH, TE, BN and SW), although the enhancements are not uniform across languages. Surprisingly, this setup shows a deterioration on the Latin languages.

Our third and final setup is Translate-EN+, which is similar to Translate-EN, but keeps the template in English for all sentences instead of translating it. Stabilizing the template improved results significantly in some languages such as French, Spanish and Russian, and gave comparable scores to Translate-EN in others.

We observe that, despite the high performance of \emph{text-davinci-003} on the translation of RU, JA and ZH, the performance on MSGM is only moderate. We hypothesize that this may be due to the fact that reasoning  tasks benefit greatly from training on programming languages, which is better represented in the top Latin languages, especially with the low proportion of multilingual data in the training data. In contrast, PaLM-540B results from \cite{mgsm} show a higher performance with the Native-CoT setup. We hypothesize that this is due to the large multilingual data proportion in its training data, which is 78\% English and 22\% for other languages \cite{mgsm}, while GPT data proportion is only 7\% non-English \cite{gpt-3}.
 
These results suggest that translation capability may not be sufficient for the model to exhibit more advanced multilingual reasoning capability, as shown by the poor performance on RU, ZH and JA. We hypothesize that the models acquire their reasoning capabilities through training on natural language multilingual data along with programming languages data, which may limit such capabilities for non-Latin and less represented languages. We think this area warrants more attention from the model developers to provide truly multilingual capabilities across a range of languages.


\section{Conclusions and Future Directions}
\label{sec:conclusion}
This work presents a comprehensive and in-depth study of the machine translation capabilities of the latest  GPT models. Our investigation covers 18 language pairs across four different domains, enabling a broad understanding of the models' general performance. We also conducted the multilingual reasoning task to examine the interaction between multilinguality and the emergent reasoning capabilities in GPT models. To provide a thorough evaluation of the models, we employed both human evaluations and the latest neural network-based automatic evaluation metrics together with the conventional machine translation evaluation metrics. In addition, we conducted extensive analysis, providing an in-depth examination of various phenomena in GPT models' translation outputs and their comparisons to state-of-the-art NMT systems.

As a result, our findings demonstrate that GPT systems can produce highly fluent and competitive translation outputs even in the zero-shot setting especially for the high-resource language translations. By utilizing the in-context learning capability of GPT models with few-shot examples, we were able to further improve translation quality. Additionally, we demonstrated that a hybrid approach, combining the latest NMT systems with GPT models, can achieve state-of-the-art translation quality.

While the use of LLMs in machine translation is a rapidly developing area, there are many research directions that can be explored to improve the quality and understanding of machine translation. Below are some of the important areas that we focus on:

\begin{itemize}
\item \textbf{Underrepresented languages}: Our study has shown that GPT models, still struggle with underrepresented languages, which makes it a critical research question to explore how to improve the translation quality for these languages.
\item \textbf{In-context learning}: GPT models have shown great potential for in-context learning, which can be leveraged to generate different styles or nuanced translations. Future research can explore how to better utilize this capability to improve translation quality.
\item \textbf{Model fusion}: The use of large-scale LLMs like GPT can be computationally expensive, and therefore, exploring how to more efficiently utilize them is an important research question. We are investigating more sophisticated fusion techniques that can achieve higher quality and efficiency.
\item \textbf{Better metrics}: The limitations of lexical matching metrics can mislead translation quality assessment. Therefore, developing metrics that can measure the contextual correctness of LLM-generated translations is essential. Future research can also explore new ways to evaluate the quality of machine translations more accurately, especially when using LLMs.
\end{itemize}

Overall, our study provides valuable insights into the strengths and weaknesses of GPT models for machine translation and opens up opportunities for future improvements and developments in this field. We investigated how GPT models can transform machine translation as it is doing with other generative tasks. We demonstrated that these models excel at translating well-represented languages in their training data, but they face challenges with less-resourced languages. We also assessed translation and reasoning tasks and detected discrepancies in the level of support of the tasks even for the same languages. One of the main benefits of training such costly models is to achieve high performance across diverse tasks and languages, but this demands more data across languages. Which poses several challenges for models scalability, diversity, and fairness. As a future research direction, we propose to tackle the challenge of enabling truly multilingual capability for such models that would enable the same capabilities across languages.

\section*{Limitations}

We conducted our evaluation on 18 translation directions with reliable test sets and baselines. The conclusion of the study should be taken in this context and not generalized to other languages without further evaluation. While a more comprehensive evaluation is needed on more languages, we should be cautious about drawing conclusions from low quality testsets or weaker baselines which are usually dominating the research results for low resource languages.

One of the limitations of this study is the inadequacy of current automatic evaluation metrics to capture the quality of GPT outputs accurately. We found that lexical comparison based metrics such as BLEU or chrF gave misleading signals, and that document-level evaluation had limited capability to realize the effect of context-based translation. These limitations stem from the inherent challenges of evaluating natural language generation systems, especially for complex tasks such as machine translation. Therefore, we complemented our automatic evaluation with a comprehensive analysis, considering all metrics together, as well as human evaluation and qualitative analysis to cover a broad range of phenomena. We recommend that readers consider the overall evaluations as a whole, rather than relying solely on a specific metric, to better understand the quality of GPT models’ machine translation capabilities.

\section*{Ethics Statement}

In our study, we evaluated the quality of GPT translations using various quantitative metrics. However, we acknowledge that these models may harbor language-specific biases and produce translations that perpetuate stereotypes and misinformation. One form of bias that we have identified is that the models perform better for some languages than others, as shown in \autoref{sec:high-resource} and \autoref{sec:low-resource}. This may create unfairness and unequal outcomes for users paying the same cost. Furthermore, the models may amplify stereotypes from the training data, leading to inaccurate translations. For instance, the model may fail to correctly translate a sentence with a female name like ``Julia'' as a male. Lastly, we also observed instances of false insertions in translations (misinformation) and hallucinations, especially in E-X translation directions. We are committed to addressing these issues and mitigating biases and misinformation in our research and future work.

\bibliography{anthology,custom}
\bibliographystyle{acl_natbib}

\onecolumn

\appendix

\clearpage

\appendix
\label{sec:appendix}
\clearpage

\clearpage


\section{Prompt Templates}
\label{sec:prompt_templates}

\begin{figure*}[!h]
    \begin{verbatim}
Translate this into 1. [target language]:

[shot n source]

1. [shot n reference]

Translate this into 1. [target language]:

[input]

1.
    \end{verbatim}
\caption{Prompt template for sentence-level translation. We use the same instruction and format as recommended in the OpenAI playground for the default sentence-level translation task \footnote{https://platform.openai.com/examples/default-translate}. Keeping the prompt format the same allows us to potentially leverage the benefits of the underlying instruction finetuning protocol to the full extent.}
\label{fig:prompt_sent}
\end{figure*}

\begin{figure*}[!h]
    \begin{verbatim}
### Translate this sentence from [source language] to [target language], Source:

[source sentence]

### Target:

    \end{verbatim}
\caption{Zero-shot prompt template for ChatGPT.}
\label{fig:prompts_chatgpt}
\end{figure*}

\begin{figure*}[!h]
    \begin{verbatim}
Document:
[shot 1 source]

[shot 2 source]

[shot n source]

####

Translate each line in document into [target language].

Translated Document:

[shot 1 reference]

[shot 2 reference]

[shot n reference]

####

Document:
[sentence 1 from context window]

[sentence n from context window]

####

Translate each line in document into [target language].

Translated Document:
    \end{verbatim}
\caption{Prompt template for document translation.}
\label{fig:prompt_doc}
\end{figure*}

%
%
%
%
%
%
%
%
%
%
%
%
%
%
%
%
%
%
%
%
%

\clearpage

\section{Few-shot Example Selection Data Pool}

\begin{table}[!htbp]
\centering
\begin{tabular}{c|c|c lc}
\hline
\multirow{2}{*}{\textbf{Language}} & \multicolumn{2}{c}{\textbf{\# of sentences}} \\
\cline{2-3}
&\textbf{Raw} & \textbf{Cleaned} \\
\hline
CS-EN & 193.5M & 175.5M \\
EN-CS & 167.9M & 151.5M \\
DE-EN/EN-DE & 295.8M & 289.1M \\ 
IS-EN/EN-IS & 4.4M & 3.7M \\
JA-EN/EN-JA & 33.9M & 33.1M \\
ZH-EN & 55.2M & 50.4M \\ 
EN-ZH & 35.5M & 31.2M \\ 
UK-EN/EN-UK & 50.6M & 45.5M \\ 
RU-EN/EN-RU & 75.0M & 65.7M \\ 
HA-EN/EN-HA & 7.2M & 727K \\ 
FR-DE/DE-FR & 17.7M & 15.6M \\ 
\hline
\end{tabular}
\caption{Size of the data pool for the few-shot example selections for each translation direction. Raw column shows the size of the original dataset which is from the WMT training dataset and Cleaned column shows high quality data after the cleaning from the original dataset.}
\label{tab:datasets-train}
\end{table}

\clearpage

\section{Human Evaluation details}

\begin{table*}[!htbp]
\centering
\vspace{0.8em}
\scalebox{0.9}{
\begin{tabular}{c|c c c c c c }
\hline

LP & WMT-HE & GPT-HE & Delta-HE & WMT-COMETkiwi & GPT-COMETkiwi & Delta COMETkiwi  \\ \hline
DE-EN & 93.8 & 92.0 & -1.8 & 81.4 & 81.4 & 0.0  \\
CS-EN & 84.1 & 85.7 & 1.7 & 82.5 & 82.5 & 0.0  \\
JP-EN & 76.5 & 74.7 & -1.8 & 80.3 & 80.8 & 0.5  \\
ZH-EN & 77.4 & 80.0 & 2.5 & 77.7 & 78.5 & 0.8  \\
RU-EN & 88.0 & 87.2 & -0.8 & 81.7 & 82.9 & 1.2  \\
UK-EN & 83.3 & 80.1 & -3.1 & 81.5 & 80.3 & -1.2  \\
IS-EN & 91.4 & 86.8 & -4.6 & 81.4 & 80.2 & -1.2  \\
HA-EN & 84.8 & 75.9 & -8.9 & 74.5 & 68.5 & -6.0  \\\hdashline
EN-DE & 95.0 & 93.4 & -1.5 & 83.6 & 82.9 & -0.7  \\
EN-CS & 85.9 & 80.5 & -5.4 & 84.2 & 83.3 & -0.9  \\
EN-JP & 79.3 & 76.0 & -3.3 & 85.8 & 85.3 & -0.5  \\
EN-ZH & 81.6 & 81.9 & 0.4 & 82.0 & 82.0 & 0.0  \\
EN-RU & 88.6 & 83.7 & -4.9 & 84.4 & 82.2 & -2.2  \\
EN-UK & 85.2 & 76.9 & -8.3 & 83.4 & 80.6 & -2.8  \\
EN-IS & 92.1 & 69.2 & -22.9 & 81.8 & 74.1 & -7.7  \\
EN-HA & 80.2 & 55.5 & -24.7 & 61.5 & 58.5 & -3.0  \\
FR-DE & 85.7 & 84.6 & -1.1 & 80.7 & 80.2 & -0.5  \\
DE-FR & 85.0 & 86.5 & 1.5 & 79.5 & 80.7 & 1.2  \\\hline

\end{tabular}}
\caption{Human  Evaluation and COMETkiwi  Results of WMT Best Systems and  GPT in details.}
\label{tab:result.he-qe}
\vspace{-1.5em}
\end{table*}

\subsection{Human Evaluation Analysis}

As shown in the examples for the first characteristic in Table~\ref{tab:he-en-ja-analysis}, GPT is good at handling misspelled words or unclosed quotes and ends translations without missing any semantic information. Also, GPT translations are reasonable for partial or incomplete colloquial source sentences whereas WMT-Best translations have extra or missing content (the third characteristic). 

On the other hand, GPT tends to generate unnatural translations for sentences with unusual or complex expressions (the second characteristic). GPT also generates hallucinated words in its translations as in the fourth characteristic. For instance, the first GPT example translation uses ``udon'' (``noodle'' in English) although the source never mentions it. In the second GPT example translation, the word ``PM'' (Prime Minister) is translated into a non-existing Japanese word that sound like an existing jargon. Unlike oscillatory hallucinations (i.e., repeating the same words or phrases many times) \cite{naacl_hallucinations}, these subtle, word-level hallucinations are problematic as they are hard to be detected and may not severely affect metric scores \cite{emnlp_long_tail_errors}.

\begin{table*}[!htbp]
    \centering
    \includegraphics[width=0.9\textwidth]{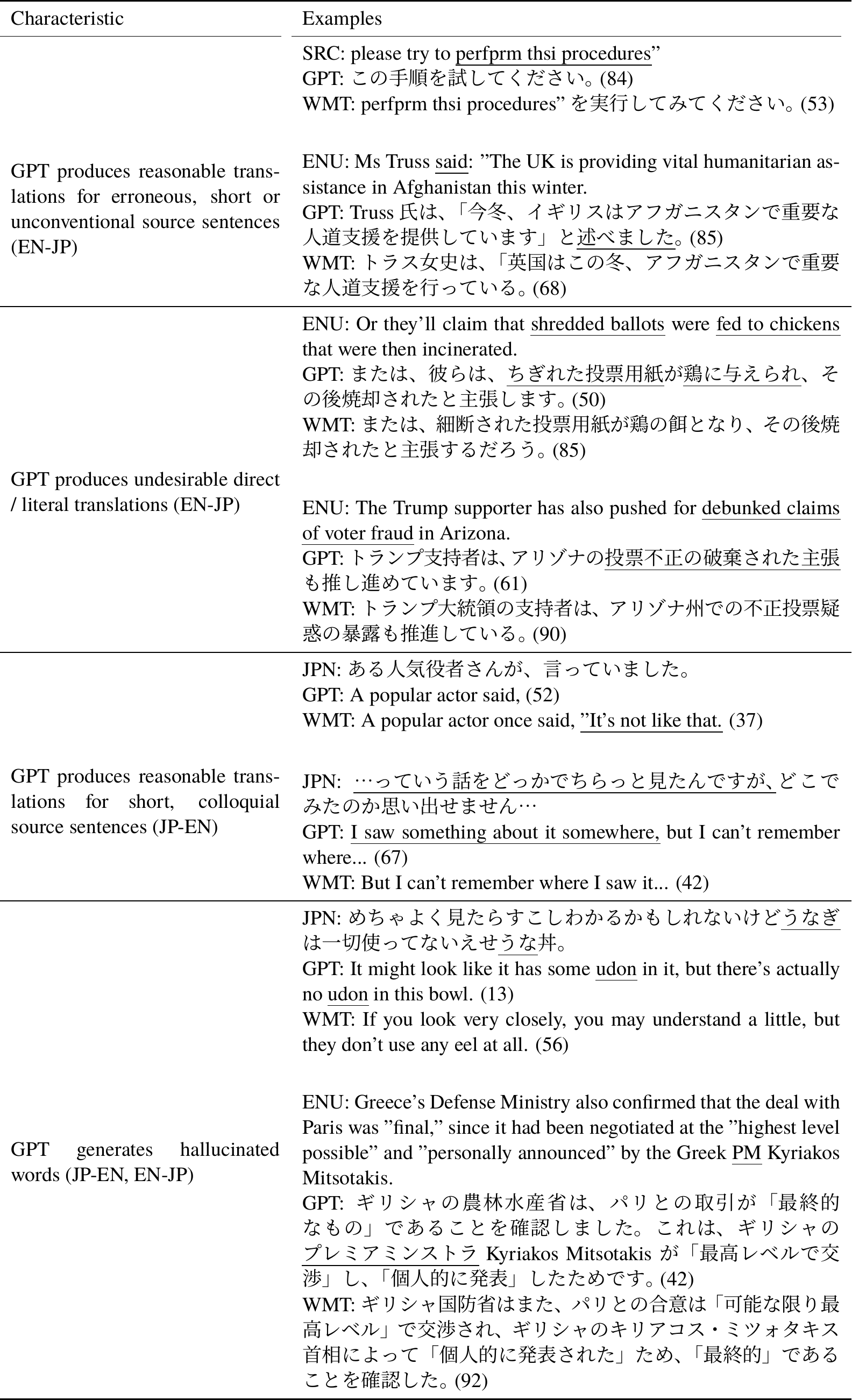}
    \caption{Qualitative analysis for GPT English from/to Japanese translations. These examples show some positive and negative characteristics indicated by human item scores (numbers in parentheses).}
    \label{tab:he-en-ja-analysis}
\end{table*}

\begin{comment}
(HM) I used content below for Table~\ref{tab:he-en-ja-analysis} 
this should be compiled with LuaLaTeX for good-looking Japanese fonts

\documentclass[11pt]{standalone}
\usepackage{luatexja}
\usepackage{fontspec}
\usepackage{graphicx} 
\usepackage{booktabs}
\usepackage{multirow}
\usepackage{ulem}
\setmainfont{TeX Gyre Termes}

\begin{document}
\begin{tabular}{p{2in}p{4in}}
    \toprule
      Characteristic &  Examples \\
    \cmidrule(r){1-1} \cmidrule(r){2-2}
    GPT produces reasonable translations for erroneous, short or unconventional source sentences (EN-JP) &
    \begin{tabular}{@{}p{4in}@{}}
    SRC: please try to \uline{perfprm thsi procedures}" \\
    GPT: この手順を試してください。 (84) \\ 
    WMT: perfprm thsi procedures"を実行してみてください。 (53) \\  \vspace{2ex}
    ENU: Ms Truss \uline{said}: "The UK is providing vital humanitarian assistance in Afghanistan this winter. \\
    GPT: Truss氏は、「今冬、イギリスはアフガニスタンで重要な人道支援を提供しています」と\uline{述べました}。 (85) \\
    WMT: トラス女史は、「英国はこの冬、アフガニスタンで重要な人道支援を行っている。 (68)    \end{tabular} \\ \midrule
    GPT produces undesirable direct / literal translations (EN-JP) &
    \begin{tabular}{@{}p{4in}@{}}
    ENU: Or they'll claim that \uline{shredded ballots} were \uline{fed to chickens} that were then incinerated. \\
    GPT: または、彼らは、\uline{ちぎれた投票用紙}が\uline{鶏に与えられ}、その後焼却されたと主張します。 (50) \\
    WMT: または、細断された投票用紙が鶏の餌となり、その後焼却されたと主張するだろう。 (85) \\ \vspace{2ex}
    ENU: The Trump supporter has also pushed for \uline{debunked claims of voter fraud} in Arizona. \\
    GPT: トランプ支持者は、アリゾナの\uline{投票不正の破棄された主張}も推し進めています。 (61) \\
    WMT: トランプ大統領の支持者は、アリゾナ州での不正投票疑惑の暴露も推進している。 (90) \\
    \end{tabular} \\ \midrule
    GPT produces reasonable translations for short, colloquial source sentences (JP-EN) &
    \begin{tabular}{@{}p{4in}@{}}    
    JPN: ある人気役者さんが、言っていました。\\
    GPT: A popular actor said, (52) \\
    WMT: A popular actor once said, \uline{"It's not like that.} (37) \\ \vspace{2ex}    
    JPN: \uline{…っていう話をどっかでちらっと見たんですが、}どこでみたのか思い出せません… \\
    GPT: \uline{I saw something about it somewhere,} but I can't remember where... (67) \\
    WMT: But I can't remember where I saw it... (42) \\ \vspace{2ex}
    \end{tabular} \\ \midrule    
    GPT generates hallucinated words (JP-EN, EN-JP) &
    \begin{tabular}{@{}p{4in}@{}}    
    JPN: めちゃよく見たらすこしわかるかもしれないけど\uline{うなぎ}は一切使ってないえせ\uline{うな}丼。 \\
    GPT: It might look like it has some \uline{udon} in it, but there's actually no \uline{udon} in this bowl. (13) \\
    WMT: If you look very closely, you may understand a little, but they don't use any eel at all. (56) \\ \vspace{2ex}
    ENU: Greece's Defense Ministry also confirmed that the deal with Paris was "final," since it had been negotiated at the "highest level possible" and "personally announced" by the Greek \uline{PM} Kyriakos Mitsotakis. \\
    GPT: ギリシャの農林水産省は、パリとの取引が「最終的なもの」であることを確認しました。これは、ギリシャの\uline{プレミアミンストラ} Kyriakos Mitsotakis が「最高レベルで交渉」し、「個人的に発表」したためです。 (42) \\
    WMT: ギリシャ国防省はまた、パリとの合意は「可能な限り最高レベル」で交渉され、ギリシャのキリアコス・ミツォタキス首相によって「個人的に発表された」ため、「最終的」であることを確認した。 (92) \\
    \end{tabular} \\   
    \bottomrule
  \end{tabular}
\end{document}